%% file: neurips_2026.tex
\documentclass{article}

\usepackage[main, final]{neurips_2026}

\usepackage[utf8]{inputenc} % allow utf-8 input
\usepackage[T1]{fontenc}    % use 8-bit T1 fonts
\usepackage{hyperref}       % hyperlinks
\usepackage{url}            % simple URL typesetting
\usepackage{booktabs}       % professional-quality tables
\usepackage{amsfonts}       % blackboard math symbols
\usepackage{nicefrac}       % compact symbols for 1/2, etc.
\usepackage{microtype}      % microtypography
\usepackage{xcolor}         % colors
\usepackage[table]{xcolor}
\definecolor{lightgreenzh}{RGB}{0,150,0}
\usepackage{arydshln}       % dashed lines in tables

\usepackage{enumitem}
\usepackage{wrapfig}
\usepackage{graphicx}
\usepackage{amsmath}
\usepackage{xspace}
\usepackage{amssymb}
\usepackage{array}
\usepackage{multirow}
\usepackage{algorithm}
\usepackage{algorithmic}
\usepackage{amsthm}
\usepackage[most]{tcolorbox}
\usepackage{makecell}
\usepackage{booktabs}
\usepackage{caption}
\newtcolorbox{mybox3}[1]{colbacktitle=white,coltitle=black,colback=white,colframe=black,fonttitle=\bfseries,title=#1,leftupper=0.5em,rightupper=0.5em,boxrule=1.0pt}
\newtcolorbox{PromptBox}[1]{
    colback=gray!10,
    colframe=black!60,
    title=#1,
    fonttitle=\bfseries,
    breakable,
    boxrule=0.5pt,
    arc=2mm,
    fontupper=\footnotesize, % 设置内容字体为 \footnotesize
}

\newcommand{\naen}[1]{\textcolor{black}{#1}}

\newcommand{\lib}{{StateTree}\xspace}

\title{\lib: Enhancing Long-Term Dialogue Reasoning via Reinforcement Learning}

\author{%
  Naen Xu$^{1,2}$\thanks{~Equal contribution.} \quad Wanqing Cui$^{2*}$ \quad Yibo Hu$^{2}$\thanks{~Project leader.} \quad Shixin Hong$^{2}$ \\
  \textbf{Hengyu An}$^{1}$ \quad \textbf{Meiguang Jin}$^{2}$ \quad \textbf{Junfeng Ma}$^{2}$ \quad \textbf{Tianyu Du}$^{1}$\thanks{~Corresponding author.} \\
  $^{1}$Zhejiang University \quad $^{2}$Taobao \& Tmall Group of Alibaba \\
  \texttt{\{xunaen, anhengyu, zjradty\}@zju.edu.cn} \\
  \texttt{\{cuiwanqing.cwq, boxuan.hyb, hongshixin.hsx\}@taobao.com} \\
  \texttt{\{meiguang.jmg, jack.majf\}@taobao.com}
}

\begin{document}

\maketitle

\vspace{-2px}
\begin{abstract}
\input{0_abstract}
\end{abstract}

\vspace{-2px}
\section{Introduction}
\vspace{-2px}
\label{introduction}

\input{1_introduction}

\section{Related Work}
\label{related_work}
\input{2_related_work}

\section{StateTree}
\label{methods}

\input{3_methods}

\vspace{-3px}
\section{Experiments}
\vspace{-2px}
\label{experiments}

\input{4_experiments}

\section{Conclusion}
\label{conclusion}
\input{6_conclusion}

\section*{Acknowledgments}
This work was partly supported by the NSFC under No. 62402418, the ``Pioneer and Leading Goose'' R\&D Program of Zhejiang under No. 2026C02A1233 and 2025C02034, the Key R\&D Program of Ningbo under No. 2024Z115, and the Ningbo Yongjiang Talent Project.

% \section*{References}
% \bibliographystyle{plain}
\bibliographystyle{unsrt}
\bibliography{custom}

\input{7_appendix}

%%%%%%%%%%%%%%%%%%%%%%%%%%%%%%%%%%%%%%%%%%%%%%%%%%%%%%%%%%%%

% \newpage
% \input{checklist.tex}

\end{document}

%% file: 0_abstract.tex
\setcounter{footnote}{0}
Large language models deployed as personalized assistants must reason over long, evolving interaction histories. However, in long-term dialogue reasoning, relevant evidence is scattered across sessions, preferences may be revised over time, and standard long-context training fails to address these challenges under data scarcity and prohibitive computational costs.
We propose StateTree, a data-driven RL method that constructs a challenging auxiliary task from scarce dialogues with verifiable ground truth. 
\lib augments multi-session dialogues with a tree-structured path-tracing task: key-value records are embedded across sessions to form a binary tree. Solving the task requires the model to traverse from root to leaf by retrieving records across sessions and comparing timestamps to resolve branches, then recover the hidden target question among distractor leaves.
We apply curriculum RL training progressively increasing tree depth and introduce a compositional variant whose edges carry step-level reasoning fragments, training the model to compose partial cues into coherent queries.
Trained on 10K-token contexts, \lib generalizes to 128K tokens without full-length RL costs and exhibits capabilities including cross-session retrieval, temporal reasoning, knowledge update, and compositional multi-hop reasoning. 
StateTree outperforms both SFT and RL-based baselines while preserving short-context general reasoning. \lib-7B achieves gains up to +23.60\% on LongMemEval (128k), and \lib-14B reaches 59.00\% accuracy on LongMemEval, surpassing QwenLong-L1-32B (45.20\%)\footnote{~Code and dataset are available at \url{https://github.com/bluedream02/StateTree}.}.

% Large language models deployed as personalized assistants must reason over long, evolving interaction histories. In long-term dialogue reasoning, however, relevant evidence is scattered across sessions, preferences may be revised over time, and standard long-context training fails under data scarcity and prohibitive computational costs.
% We propose StateTree, a data-driven RL pseudo-task that constructs challenging training signals from scarce dialogues with verifiable ground truth. \lib embeds key-value records across multi-session dialogues to form a binary tree; the model must traverse from root to leaf via cross-session retrieval and temporal comparison to recover the hidden target question.
% A structured curriculum progressively increases tree depth and introduces a compositional variant whose edges carry step-level reasoning fragments that the model must aggregate into coherent queries.
% Trained on only 616 examples at 10K tokens, \lib generalizes to 128K tokens without full-length RL costs, eliciting targeted behaviors including cross-session retrieval, temporal reasoning, knowledge update, and compositional multi-hop reasoning while preserving short-context abilities. 
% StateTree outperforms both SFT and RL baselines: on the 128K-token LongMemEval benchmark, \lib-7B improves by +23.60\% and \lib-14B reaches 59\% accuracy (+13.8\%), surpassing QwenLong-L1-32B (45.20\%).

%% file: 1_introduction.tex
% Large Language Models (LLMs) have demonstrated impressive capabilities across diverse tasks \citep{srivastava2023beyond, zhou2023instruction}, serving as personalized assistants for writing \citep{mysore-etal-2024-pearl,tian-etal-2024-large-language}, recommendation \citep{hua2023tutorial}, and consultation \citep{mysore-etal-2024-pearl, xie2024travelplanner}. Increasingly, these agents are expected to support long-term, multi-session interactions \citep{du2025bridging, du2026memoryt}.

Large Language Models (LLMs) are increasingly deployed as personalized assistants for writing \citep{mysore-etal-2024-pearl,tian-etal-2024-large-language}, recommendation \citep{hua2023tutorial,li2025multi,11270220,chen2024post}, and consultation \citep{mysore-etal-2024-pearl, xie2024travelplanner}, while supporting long-term, multi-session dialogues \citep{du2025bridging}. 
However, LLMs struggle to leverage personal knowledge accumulated over extended interaction histories \citep{zhong2024memorybank, laban2026llms}, leading to degraded accuracy and reduced user satisfaction \citep{ling2025longreason}. While memory-augmented systems \citep{zhong2024memorybank, fang2026lightmem} improve personalization via compression, indexing, and retrieval over chat histories, they still rely on the backbone LLM's reasoning ability. Meanwhile, most long-context research targets static documents~\citep{yang2018hotpotqa,ho2020constructing}, treating dialogue history as flat text and failing to adapt to evolving user personas \citep{maharana2024evaluating, wu2025longmemeval}. Long-term dialogue reasoning presents three challenges:

\begin{itemize}[nosep,leftmargin=11pt,topsep=0pt]

    \item \textbf{Data Scarcity ($\mathbb{C}1$):} 
    % Effective training requires challenging, multi-session interaction histories that trigger multi-hop reasoning. 
    % Existing datasets such as LoCoMo~\citep{maharana2024evaluating} contain only 10 dialogues ($\sim$616 QA pairs). Using LLMs to synthesize more complex dialogue datasets poses challenges because we cannot ensure the answers correctness.
    Existing multi-session datasets such as LoCoMo~\citep{maharana2024evaluating} contain only 10 dialogues (1540 QA pairs), and LLM-synthesized dialogues cannot guarantee answer correctness.
    
    \item \textbf{Complex Dialogue Reasoning ($\mathbb{C}2$):} Unlike static documents, long-term dialogues are non-stationary: evidence is scattered across sessions with timestamps, and preferences may be revised later. Answering question requires chaining multiple pieces of evidence, yet LLMs frequently fail at cross-session retrieval, temporal reasoning, knowledge update, and multi-hop composition.

    \item \textbf{Prohibitive Computational Cost ($\mathbb{C}3$):} % Achieving strong long-term performance typically requires training at near-target lengths \citep{liu2024deepseek,li2025minimax}, but long contexts incur prohibitive compute and memory costs, making direct training infeasible at standard compute scales. Furthermore, exclusive long-context training risks degrading essential short-context and general reasoning abilities \citep{peng2023yarn,longrope2}.
    Training at near-target context lengths \citep{liu2024deepseek,li2025minimax} incurs prohibitive compute costs and risks degrading short-context and general reasoning abilities \citep{peng2023yarn,longrope2}.

\end{itemize}

% To mitigate data scarcity ($\mathbb{C}1$), constructing synthetic trajectories for Supervised Fine-Tuning (SFT) is a common strategy \citep{prabhakar2025apigen,liu2025prorl}. However, LLMs trained on these static traces show limited improvement and often fail to generalize to out-of-distribution datasets. \naen{While Reinforcement Learning (RL) \citep{o1, guo2025deepseek,shao2024deepseekmath} offers an alternative to elicit reasoning through exploration \citep{chu2025sft},} its effectiveness in long-term dialogue is fundamentally constrained by the lack of challenging, task-specific training data ($\mathbb{C}1$) and the instability from prohibitive costs of long interaction chains ($\mathbb{C}3$).

While synthetic trajectories for SFT \citep{prabhakar2025apigen,liu2025prorl,chu2025sft} mitigate data scarcity ($\mathbb{C}1$), models trained on static traces show limited improvement and often fail to generalize to out-of-distribution datasets. Reinforcement Learning (RL) \citep{guo2025deepseek,shao2024deepseekmath} offers an alternative, yet its effectiveness in long-term dialogue is constrained by the lack of challenging training data with ground truth ($\mathbb{C}1$) and prohibitive costs of long interaction chains ($\mathbb{C}3$).
This leads to our central question: \textit{How can we elicit robust \textbf{long-term dialogue reasoning} ($\mathbb{C}2$) from extremely \textbf{scarce data} ($\mathbb{C}1$) without \textbf{prohibitive computational overhead} of full-context RL training ($\mathbb{C}3$)?}

\begin{figure}[!t]
    \centering
    % \vspace{-1mm}
    \includegraphics[width=\textwidth]{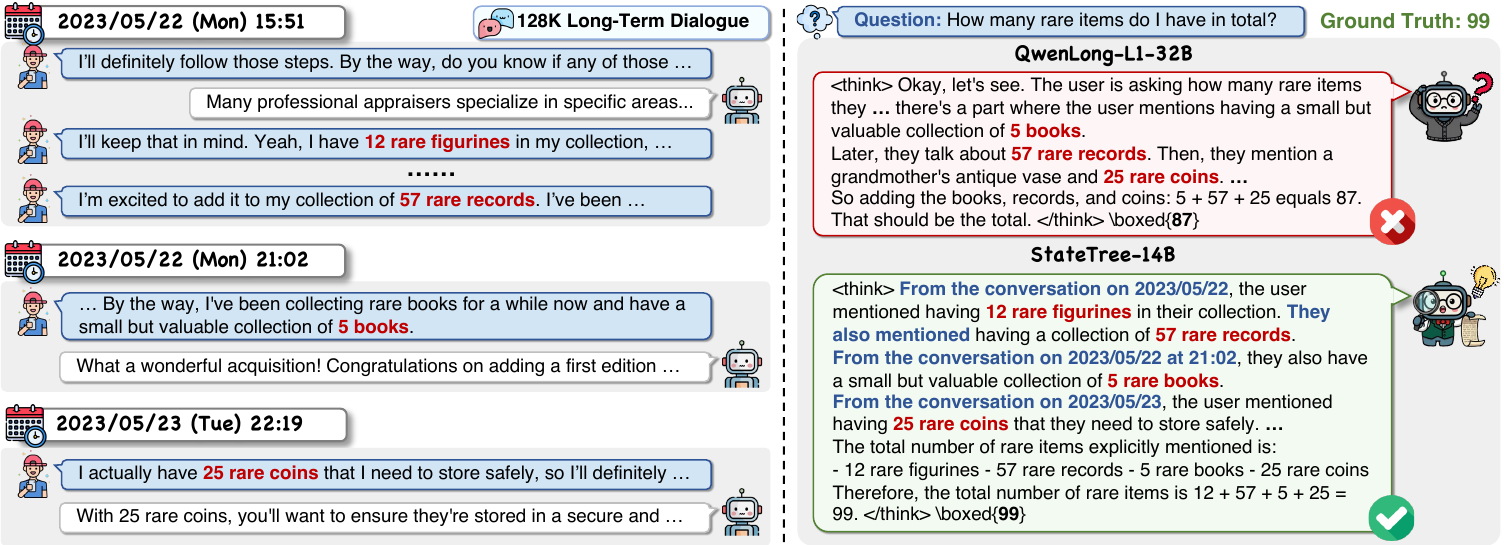}
    % \vspace{-1mm}
    % \vspace{-15px}
    \caption{Model trajectories in multi-session dialogue reasoning. (i) \naen{QwenLong-L1-32B gets lost across sessions and misses information.} (ii) The \lib-trained model shows cross-session retrieval, temporal reasoning, knowledge update, and multi-hop reasoning, improving reasoning reliability.}

    \label{fig:examples}
    \vspace{-20px}
    % \vspace{-5mm}
\end{figure}

% In response, we introduce \textbf{\lib}, a data-driven RL framework that elicits long-term dialogue reasoning.
% Our key insight is that the reasoning patterns required for multi-session dialogues can be taught through a carefully designed \emph{synthetic task} embedded within authentic dialogues.
% In response, we introduce \textbf{\lib}, \naen{a data-driven RL pseudo-task that elicits long-term dialogue reasoning by constructing challenging training signals from scarce data with verifiable ground truth, without prohibitive training overhead.

In response, we introduce \textbf{\lib}, \naen{a data-driven RL pseudo-task that constructs challenging synthetic auxiliary tasks embedded within authentic dialogues. % , targeting the capabilities required for long-term dialogue reasoning and providing verifiable ground truth.
Our key insight is that multi-session dialogue reasoning can be reframed as a path-search problem: relevant evidence forms a navigable structure with temporal constraints and hierarchical dependencies. These patterns can be taught through structured synthetic tasks that provide verifiable ground truth.}
Concretely, \naen{\lib embeds key-value records into long-term dialogues, transforming limited data into challenging training instances ($\mathbb{C}1$).
The same key is placed across multiple sessions with different timestamps, and the linked records form a binary tree. The model must trace a path from root to leaf by comparing timestamps and following the most recent record at each branch point, until the leaf reveals the target question ($\mathbb{C}2$).}
% augments multi-session transcripts with a \emph{StateTree} task ($\mathbb{C}1$): key-value records are scattered across sessions to form a binary tree, and the model must navigate from a root key to a leaf to recover the hidden question before answering it. 
We apply a curriculum RL strategy that progressively increases tree depth and reasoning complexity; its final stage introduces a \emph{Compositional StateTree} whose edges carry step-level reasoning fragments that the model must aggregate into a coherent query.
% By training at only 10K context on 616 examples, \lib avoids the prohibitive cost of full-length RL ($\mathbb{C}3$).
By training at contexts of 10K tokens, \lib avoids the prohibitive cost of full-length RL ($\mathbb{C}3$).
% We design a \emph{State Tree} auxiliary task that augments authentic multi-session transcripts with UUID-keyed records forming a tree structure, so that solving the task requires the model to acquire specific reasoning patterns critical for long-term interaction understanding.
% Starting from a root node, \lib inserts competing records with the same UUID key across different sessions, each pointing to the next hop (or to leaf questions). At each fork, the same key appears in multiple sessions, and the model must select the record from the most recent session---a temporal discrimination signal. In the most advanced stage, edges carry semantic step fragments that the model must aggregate to recover the hidden question. To solve this, a model must: (\textit{i}) retrieve matching records scattered across sessions (\emph{cross-session retrieval}), (\textit{ii}) compare session timestamps and select the most recent record (\emph{temporal reasoning} and \emph{knowledge update}), and (\textit{iii}) chain multiple hops and compose the collected evidence into a coherent answer (\emph{multi-hop reasoning}). 
% Solving these tasks forces the model to move beyond retrieval toward multi-hop arbitration and grounded reasoning. 
% We employ Group Relative Policy Optimization (GRPO) with a token-level F1 reward that accommodates linguistic variation while mitigating reward hacking.

Our experiments show that \lib substantially improves Qwen2.5-7B-Instruct, Qwen2.5-14B-Instruct and Qwen3-8B on long-term dialogue reasoning. 
\lib-14B achieves an average accuracy of 60.91\% on LoCoMo, \naen{surpassing much larger baselines} including QwenLong-L1-32B. \naen{Trained on contexts of 10K tokens, \lib models generalize to long-term conversation reasoning on much longer 128K benchmarks (LongMemEval, PersonaMem-128k), and \lib preserves short-context reasoning abilities.}
Notably, a qualitative comparison reveals that \lib-trained models exhibit four emergent reasoning behaviors (cross-session retrieval, temporal reasoning, knowledge update, and multi-hop composition) that directly align with the four challenges encoded in the StateTree design, confirming that the synthetic task successfully imparts targeted reasoning patterns that generalize beyond the training distribution.
Our main contributions are:

% \begin{itemize}[nosep,leftmargin=11pt,topsep=0pt]
% \item We propose \lib, an RL framework that embeds a \emph{StateTree} task within authentic multi-session dialogues to train cross-session retrieval, temporal reasoning, knowledge update, and compositional multi-hop reasoning.
% \item We develop a structured curriculum that incrementally increases tree depth and introduces semantic decomposition, enabling stable RL training with rich learning signals from only $\sim$616 examples.
% \item \lib generalizes from 16K to 128K tokens, achieves competitive performance with much larger models while preserving general reasoning, and the emergent reasoning behaviors align with the StateTree design.
% \end{itemize}

\begin{itemize}[nosep,leftmargin=11pt,topsep=0pt]
% \item \naen{We propose \lib, a data-driven RL pseudo-task that embeds tree-structured auxiliary tasks within authentic multi-session transcripts to construct challenging training signals from scarce data with verifiable ground truth, targeting cross-session retrieval, temporal reasoning, knowledge update, and compositional multi-hop reasoning.}
% \item We develop a structured curriculum that incrementally increases tree depth and introduces semantic decomposition, enabling stable RL training with rich learning signals from only $\sim$616 examples.

\item \naen{We propose \lib, a data-driven RL pseudo-task that augments scarce multi-session dialogues with tree-structured key-value records, transforming limited data into challenging training instances with verifiable ground truth to enhance long-term dialogue reasoning.}

\item \naen{We develop a curriculum RL strategy that progressively increases tree depth and introduces a compositional variant with step-level semantic fragments, enabling stable training and eliciting targeted reasoning behaviors.}

% \item Trained on only 616 examples, \lib elicits emergent reasoning behaviors that directly correspond to the task design, generalizes from 10K to 128K tokens, and preserves short-context abilities while matching the performance of much larger models.
\item StateTree elicits emergent reasoning behaviors correspond to its task design. Despite training on only 616 examples at 10K tokens, it generalizes to 128K contexts without full-length RL costs, preserves short-context abilities, and matches or exceeds the performance of much larger models.

\end{itemize}

%% file: 2_related_work.tex
\textbf{Evaluating Long-term Multi-Session Dialogues.}
As LLMs are increasingly deployed as personalized assistants, understanding their ability to reason over extended interaction histories has become critical.
A key finding is that models exhibit a ``lost-in-the-middle'' effect, with greater difficulty recalling information in the middle of the context~\citep{liu2024lost}. 
\cite{laban2026llms} further show that when LLMs take a wrong turn in multi-turn dialogue, they fail to recover.
Recently, evaluation has shifted toward more realistic dialogue interactions with sessions ranging from 10K to over 100K tokens.
Representative benchmarks include LoCoMo~\cite{maharana2024evaluating}, LongMemEval~\cite{wu2025longmemeval}, and PersonaMem~\cite{jiang2025know}. These evaluations reveal that even strong models struggle with long-term dialogue reasoning, motivating two lines of work: memory-augmented systems~\cite{xu2026when} and long-context reasoning methods.

\textbf{Long-Context Reasoning.}
Memory-augmented systems \cite{zhong2024memorybank, fang2026lightmem} externalize memory through compression, indexing, and retrieval, but ultimately depend on the backbone LLM's ability to reason over retrieved context. Advanced long-context reasoning primarily utilizes synthetic-data SFT \citep{li2024large, li2024making} and RL \citep{guo2025deepseek}, both of which are constrained by the biases of static synthetic traces or the base model's intrinsic reasoning capacity. 
For example, \cite{li2024large} proposes SEALONG, a self-improvement approach that samples multiple reasoning outputs and selects high-quality traces for SFT or preference alignment. 
QwenLong-L1 \citep{wan2025qwenlong} applies RL to extend long reasoning trajectories up to 60K tokens. 
Inspired by needle-in-a-haystack~\citep{needlehaystack}, which measures models' retrieval ability in extremely long-text settings, LoongRL~\cite{wang2026loongrl} inserts chains that hide the true question among distracting documents to support advanced long-context reasoning.
While these approaches enhance long-context reasoning, they are designed for static documents rather than the non-stationary, multi-session dialogue setting.

%% file: 3_methods.tex
\begin{figure}[!t]
    \centering
    % \vspace{-1mm}
    \includegraphics[width=\textwidth]{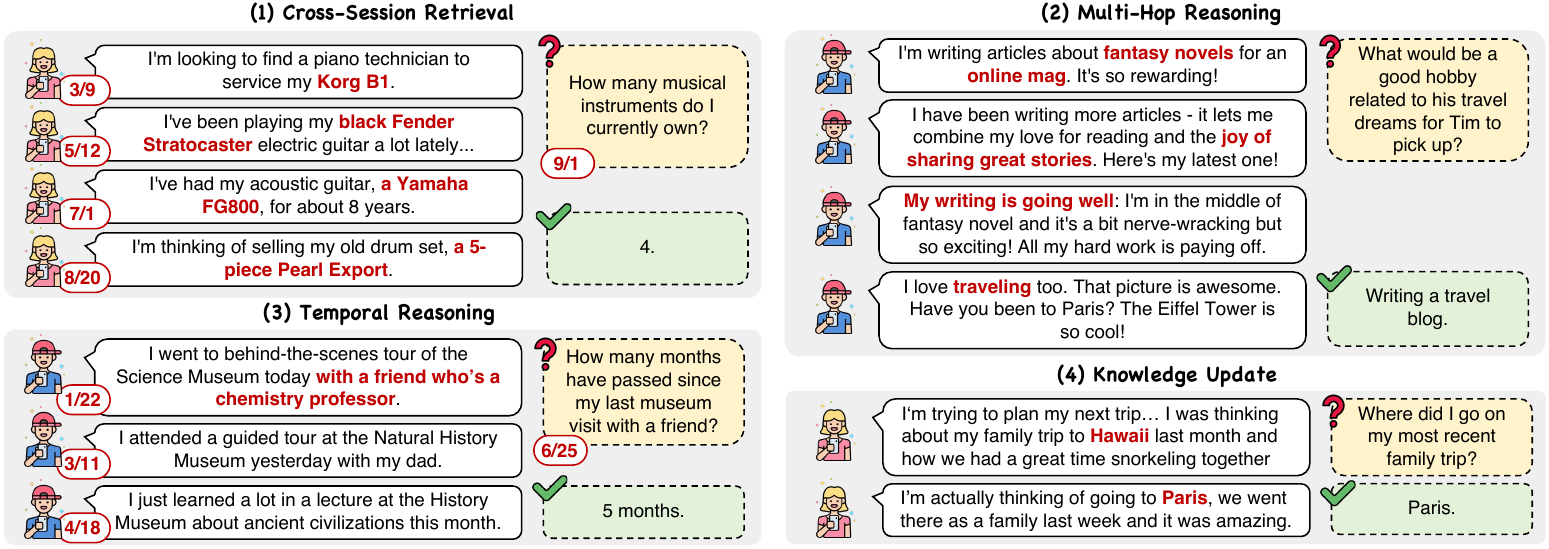}
    % \vspace{-1mm}
    \caption{Examples of the challenges in multi-turn dialogue. For each example, we show the associated evidence statements on the left and the question with the answer on the right.}
    \label{fig:challenges}
    \vspace{-10px}
    % \vspace{-5mm}
\end{figure}

As shown in Figure \ref{fig:challenges}, long-term interaction dialogues pose four challenges: 
(\textit{i}) \emph{\textbf{Cross-Session Retrieval}}, \naen{which requires precisely locating sparse, relevant evidence scattered across numerous sessions} \citep{maharana2024evaluating}; 
(\textit{ii}) \emph{\textbf{Multi-Hop Reasoning}}, \naen{necessitating bridging multiple retrieval steps across sessions and synthesizing the aggregated fragments into a final answer} \citep{maharana2024evaluating, wu2025longmemeval};
(\textit{iii}) \emph{\textbf{Temporal Reasoning}}, \naen{which involves comparing timestamps across sessions to ground events in chronology} \citep{wu2025longmemeval}; 
and (\textit{iv})  \emph{\textbf{Knowledge Update}}, \naen{which requires recognizing that the same topic may be discussed at different times and that the most recent information should take precedence} \citep{dean2022preference}.

\begin{figure}[!t]
    \centering
    % \vspace{-1mm}
    \includegraphics[width=\textwidth]{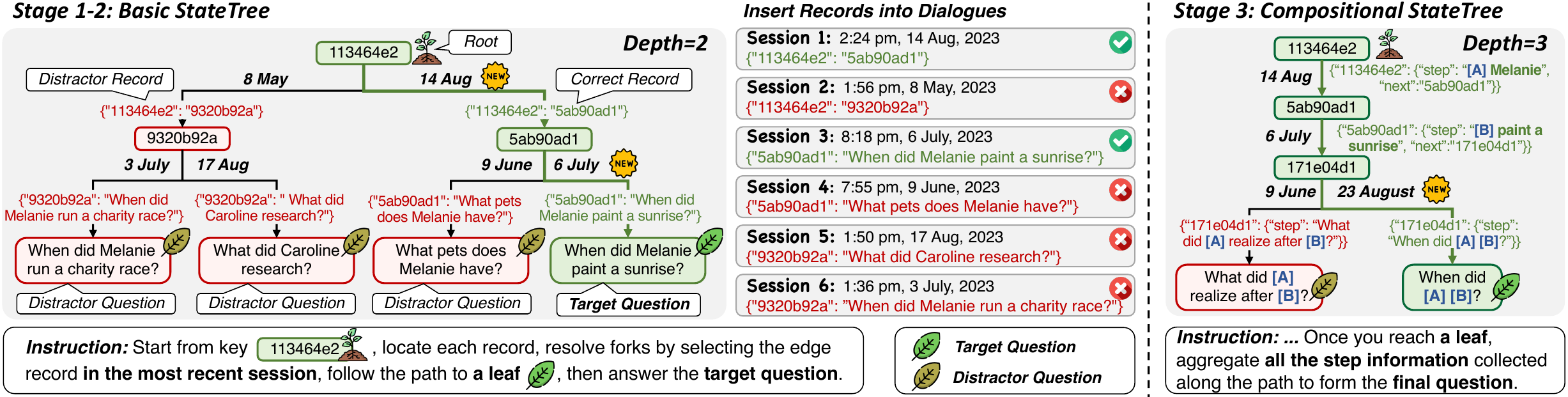}
    % \vspace{-1mm}
    \caption{The \lib data construction pipeline. \textbf{Left:} Basic StateTree embeds key-value records across sessions into a tree; the model traverses root-to-leaf to recover the target question. \textbf{Right:} Compositional StateTree augments edges with step-level fragments aggregates into the final question.}
    \vspace{-12px}
    
    \label{fig:data_construction}
    % \vspace{-5mm}
\end{figure}

% To overcome these challenges, we propose \lib, a \naen{data-driven RL pseudo-task} that elicits structured reasoning patterns in LLMs for long-term dialogue understanding (see Figure~\ref{fig:data_construction}). 
To overcome these challenges, we propose \lib, a \naen{data-driven RL pseudo-task that constructs challenging synthetic auxiliary tasks embedded within authentic dialogues, targeting the capabilities required for long-term dialogue understanding and providing verifiable ground truth} (see Figure~\ref{fig:data_construction}). 
It combines (i) a data construction pipeline (Section~\ref{sec:data_construction_pipeline}) for challenging task generation;
and (ii) GRPO with structured curriculum RL training for progressive skill acquisition through increasing tree depth and reasoning complexity, together with a combined reward function using exact match reward to provide clear optimization signals that directly target answer correctness (Section~\ref{sec:rl}). 
These pillars collectively provide a manageable learning path for stable, data-efficient LLM training.
Table~\ref{tab:challenge_mapping} summarizes how each challenge is addressed by a specific design choice in \lib.

\vspace{-3px}
\subsection{Data Construction Pipeline}
\label{sec:data_construction_pipeline}

The core challenge in training LLMs for long-term dialogue reasoning is the absence of tasks that demand multi-step reasoning over extended dialogue histories. To address this, we design a \naen{data-driven RL pseudo-task} with a data construction pipeline that embeds enhanced reasoning challenges directly within authentic multi-session dialogue data.

\begin{table}[t]
    \centering
    \caption{Mapping between multi-turn dialogue challenges and \lib design choices.}
    \label{tab:challenge_mapping}
    \resizebox{\textwidth}{!}{
    \begin{tabular}{ll}
        \toprule
        \textbf{Challenge} & \textbf{Design in \lib} \\
        \midrule
        Cross-Session Retrieval & Edge records distributed evenly across $S$ sessions, preventing positional shortcuts \\
        Multi-Hop Reasoning & $D$-level tree traversal; compositional StateTree requires aggregating step fragments into the question \\
        
        Temporal Reasoning & \naen{At each fork, compare session timestamps to determine chronological ordering}  \\
        Knowledge Update &  \naen{Correct edge is placed in a newer session than the distractor, mirroring real recency preference} \\

        \bottomrule
    \end{tabular}
    }
\end{table}

\textbf{Base Data.}
We build upon multi-session dialogue datasets (e.g., LoCoMo~\cite{maharana2024evaluating}) where each sample consists of multiple dialogue sessions between two speakers, each annotated with a timestamp. Each sample is paired with factual QA pairs whose answers can be derived from the dialogue content.

\textbf{Task Overview.}
% At the heart of our data construction is a \emph{StateTree}---a complete binary tree of depth $D$ whose edges are encoded as JSON key-value records and scattered across the dialogue sessions. To solve the task, the model must traverse the tree by locating these \emph{edge records} across sessions, resolving forks using temporal cues, and arriving at a leaf that contains a hidden question.
% \naen{The core of our data construction is \emph{StateTree}, a complete binary tree of depth $D$ built around a target question-answer pair $(q_i, a_i)$ drawn from the dataset. The target question $q_i$ is hidden in one of the $2^D$ leaf nodes, while all tree edges are encoded as JSON key-value records and scattered across the dialogue sessions. To solve the task, the model must traverse the tree from the root by locating these \emph{edge records} across sessions, resolving forks using temporal cues, arriving at the leaf that recovers the hidden question, and finally answering it based on the dialogue content to produce $a_i$.}
\naen{The core of our data construction is \emph{StateTree}, a complete binary tree of depth $D$ built around a target question-answer pair $(q_i, a_i)$. The target question $q_i$ is hidden in one leaf node, while all edges are encoded as JSON key-value records scattered across dialogue sessions. Solving the task requires traversing from root to leaf to recover the hidden question and produce $a_i$.}

Consider a multi-session dialogue $\mathcal{S} = \{s_1, \ldots, s_S\}$, where each session is annotated with a timestamp, paired with a set of questions $\mathcal{Q}$ and their ground-truth answers. For a target question $q_i \in \mathcal{Q}$ with answer $a_i$, we construct a StateTree $\mathcal{T}$ and embed its edges as JSON records throughout the sessions, yielding an augmented dialogue $\mathcal{S}'$. Each training instance is then a tuple $(\mathcal{S}', \mathcal{P}, a_i)$: the model receives $\mathcal{S}'$ together with an instruction prompt $\mathcal{P}$ specifying the root key $k_{\text{root}}$ and traversal rules, and must find $q_i$ and produce $a_i$.
The design follows three principles, each targeting a specific challenge: \textbf{(i)} the original question $q_i$ is \emph{hidden} and edge records are distributed across all $S$ sessions, so the model must perform \emph{cross-session retrieval} to locate them; \textbf{(ii)} at each fork, the correct edge record resides in a \emph{more recent session} than the distractor, forcing \emph{temporal reasoning} and \emph{knowledge update}; \textbf{(iii)} the model must navigate $D$ levels of forks to reach the leaf, then answer the recovered question, exercising \emph{multi-hop reasoning}.
We employ two tree formats: a \emph{Basic StateTree} whose leaves directly contain questions, and a \emph{Compositional StateTree} whose edges carry step-level reasoning fragments that must be aggregated to recover the question.

% \noindent

\begin{minipage}[t]{0.49\textwidth}
\vspace{-18px}
\begin{algorithm}[H]
    \small
    \captionsetup{font=small, labelfont={bf, small}}
    \caption{Basic StateTree Construction.}
    \label{alg:basic_tree}
    \begin{algorithmic}[1]
        \REQUIRE $\mathcal{S}=\{s_1,\dots,s_S\}$, QA pairs $\mathcal{Q}$, depth $D$
        \ENSURE \!Augmented dialogue $\mathcal{S}'\!$, prompt $\mathcal{P}\!$, answer $a_i\!$
        \STATE $(q_i, a_i) \sim \mathcal{Q}$
        \STATE Build $\mathcal{T}$ (depth $D$, internal nodes $V_{\text{int}}$, leaves $L$); assign UUID key $k_v$ to each node
        \STATE Sample $\ell^* \in L$; set $q_{\ell^*} \gets q_i$ and $q_\ell \sim \mathcal{Q}\setminus\{q_i\}$ for $\ell \neq \ell^*$
        \FOR{each $v \in V_{\text{int}}$ with children $u_1, u_2$}
            \FOR{$j \in \{1,2\}$}
                \IF{$u_j \in V_{\text{int}}$}
                    \STATE $r_j \gets \{k_v : k_{u_j}\}$
                \ELSE
                    \STATE $r_j \gets \{k_v : q_{u_j}\}$
                \ENDIF
            \ENDFOR
            \IF{$v$ on root-to-$\ell^*$ path}
                \STATE $u_{\text{corr}} \gets$ on-path child, $u_{\text{dist}} \gets$ sibling
                \STATE Pick $s_a, s_b \in \mathcal{S}$ with $\text{timestamp}(s_b) > \text{timestamp}(s_a)$
                \STATE Assign $r_{\text{corr}}$ to $s_b$ and $r_{\text{dist}}$ to $s_a$
            \ELSE
                \STATE Assign $r_1, r_2$ to distinct sessions evenly
            \ENDIF
        \ENDFOR
        \STATE Insert all records at sentence boundaries $\to \mathcal{S}'$
        \STATE $\mathcal{P} \gets \textsc{Prompt}(k_{\text{root}}, \text{traversal rules})$; \RETURN $(\mathcal{S}', \mathcal{P}, a_i)$
    \end{algorithmic}
\end{algorithm}
\end{minipage}
\hfill
\begin{minipage}[t]{0.49\textwidth}
\vspace{-18px}
\begin{algorithm}[H]
    \small
    \captionsetup{font=small, labelfont={bf, small}}
    \caption{Compositional StateTree Construction.}
    \label{alg:comp_tree}
    \begin{algorithmic}[1]
        \REQUIRE $\mathcal{S}=\{s_1,\dots,s_S\}$, target $(q_i, a_i)$, depth $D$, axes $\{A_1,\dots,A_D\}$
        \ENSURE \!Augmented dialogue $\mathcal{S}'\!$, prompt $\mathcal{P}\!$, answer $a_i\!$
        \STATE Build $\mathcal{T}$ (depth $D$, keys $k_v$); sample target leaf $\ell^*$
        \STATE Decompose $q_i$ via LLM along axes, yielding target step $z_{(v,u)}$ for each edge on the root-to-$\ell^*$ path %  (depth-$d$ edges map to axis $A_d$)
        \STATE Generate distractor steps for all other root-to-leaf paths via LLM, differing in $\ge 1$ axis
        \FOR{each internal node $v$ with children $u_1, u_2$}
            \FOR{$j \in \{1,2\}$}
                \IF{$u_j \notin L$}
                    \STATE $r_j \!\gets \!\{k_v \!:\! \{\texttt{step}: z_{(v,u_j)}, \texttt{next}: k_{u_j}\}\}$
                \ELSE
                    \STATE $r_j \gets \{k_v : \{\texttt{step}: z_{(v,u_j)}\}\}$
                \ENDIF
            \ENDFOR
            \IF{$v$ on root-to-$\ell^*$ path}
                \STATE $u_{\text{corr}} \gets$ on-path child, $u_{\text{dist}} \gets$ sibling
                \STATE Pick $s_a, s_b \in \mathcal{S}$ with $\text{timestamp}(s_b) > \text{timestamp}(s_a)$
                \STATE Assign $r_{\text{corr}}$ to $s_b$ and $r_{\text{dist}}$ to $s_a$
            \ELSE
                \STATE Assign $r_1, r_2$ to distinct sessions evenly
            \ENDIF
        \ENDFOR
        \STATE Insert all records at sentence boundaries $\to \mathcal{S}'$
        \STATE $\mathcal{P}\!\! \gets \!\!\textsc{Prompt}(k_{\text{root}}, \text{traversal and composition rules})$; 
        \vspace{-10px}
        \RETURN $(\mathcal{S}', \mathcal{P}, a_i)$
    \end{algorithmic}
\end{algorithm}
\end{minipage}

\vspace{10px}

\textbf{Basic StateTree.}
The Basic StateTree $\mathcal{T}$ is a complete binary tree of depth $D$.
Each node is identified by a unique \emph{key}---a randomly generated UUID string (e.g., \texttt{f391e945-...-9bbcecd4b90d}) carrying no semantic meaning, forcing the model to perform explicit lookup-and-follow operations rather than content-based shortcuts.
Each of the $2^D - 1$ internal nodes has exactly two outgoing edges, encoded as JSON key-value records (e.g., \texttt{\{``$k_i$'': ``$k_j$''\}}).
The $2^D$ leaf nodes each contain a natural-language question: exactly one holds the target question $q_i$; the rest hold distractor questions sampled from $\mathcal{Q} \setminus \{q_i\}$, ensuring all leaf questions are plausible and answerable from $\mathcal{S}'$, so the model cannot bypass tree navigation by simply selecting the most relevant-sounding question.
There exists a unique correct path $k_{\text{root}} \to k_2 \to \cdots \to k_D \to q_i$ from the root to the target leaf.
The tree contains $2(2^D - 1)$ edge records in total, all scattered across the dialogue sessions.

For each internal node on the correct path, its correct and distractor edge records are placed in \emph{different} sessions, with the correct record always in the \emph{more recent} one to force the model to \naen{perform \emph{temporal discrimination} by comparing session timestamps to identify the most recent record.} All edge records are inserted at sentence boundaries and distributed evenly across sessions to blend naturally into the dialogue flow.
Algorithm~\ref{alg:basic_tree} formalizes the construction of a Basic StateTree. The pipeline first builds a complete binary tree and scatters its edges across the dialogue sessions, then generates an instruction prompt that asks the model to traverse the tree and answer the recovered question.
An example instance is provided in Appendix~\ref{app:record_examples}.

\textbf{Compositional StateTree.}
While the Basic StateTree trains cross-session retrieval, temporal reasoning, and knowledge update, the model never needs to \emph{compose} information gathered along the path, which is a critical aspect of multi-hop reasoning.
% its leaves contain ready-made questions, so
% In multi-hop reasoning, the model needs to compose information gathered along the path.
% To train this compositional form of multi-hop reasoning, 
Thus, we introduce the \emph{Compositional StateTree}, a $D$-level binary tree whose edges carry structured \emph{step} information instead of bare key pointers.
% Concretely, given the target question $q_i$ for a dialogue between two speakers, we use an LLM to decompose $q_i$ into a $2 \times 2 \times 2$ hierarchy along three semantic axes: (i) \emph{\textbf{person selection}}, which speaker is relevant to the dialogue; (ii) \emph{\textbf{event category}}, which broad type of activity is involved; and (iii) \emph{\textbf{question specificity}}, the precise question detail. The three axes correspond directly to the three levels of the depth-$3$ binary tree with 8 leaves.
% The LLM generates alternative combinations along each axis for the remaining $7$ leaves, producing distractor questions that share partial semantic overlap with $q_i$ but differ in at least one axis (Appendix~\ref{app:decomp_prompt}).
% Each edge record is encoded as a nested JSON object \texttt{\{``$k_i$'': \{``step'': ``...'', ``next'': ``$k_j$''\}\}} for internal edges, or \texttt{\{``$k_i$'': \{``step'': ``...''\}\}} for leaf edges (no \texttt{next} field). At each level, the \texttt{step} field provides a reasoning fragment:
% \begin{itemize}[nosep,leftmargin=11pt,topsep=0pt]
%     \item \textbf{Level 0:} person identity, e.g., ``\texttt{[A] Jon}''
%     \item \textbf{Level 1:} event category, e.g., ``\texttt{[B] expanding his studio's social media presence}''
%     \item \textbf{Level 2:} question specificity, e.g., ``\texttt{When did [A] start [B]?}''
% \end{itemize}
Concretely, given the target question $q_i$ for a dialogue between two speakers, we use an LLM (prompt in Appendix~\ref{app:decomp_prompt}) to decompose $q_i$ into a $2 \times 2 \times 2$ hierarchy along three semantic axes, which define the levels of a depth-$3$ binary tree with 8 leaves.
Each edge carries a \texttt{step} field that provides a reasoning fragment at the corresponding level:
\begin{itemize}[nosep,leftmargin=11pt,topsep=0pt]
    \item \textbf{Person selection (Level 1):} which speaker is relevant to the dialogue, e.g., ``\texttt{[A] Jon}''
    \item \textbf{Event category (Level 2):} which broad type of activity is involved, e.g., ``\texttt{[B] expanding his studio's social media presence}''
    \item \textbf{Question specificity (Level 3):} the precise question detail, e.g., ``\texttt{When did [A] start [B]?}''
\end{itemize}
The LLM generates alternative combinations along each axis for the remaining 7 leaves, producing distractor questions that share partial semantic overlap with $q_i$ but differ in at least one axis.
Each edge record is encoded as a nested JSON object \texttt{\{``$k_i$'': \{``step'': ``...'', ``next'': ``$k_j$''\}\}} for internal edges, or \texttt{\{``$k_i$'': \{``step'': ``...''\}\}} for leaf edges (no \texttt{next} field). Algorithm~\ref{alg:comp_tree} formalizes the construction of a Compositional StateTree, and examples are provided in Appendix~\ref{app:record_examples}.

The model must \textbf{(i)} navigate the tree using the same temporal discrimination as the Basic StateTree; \textbf{(ii)} accumulate the \texttt{step} fragments along the correct root-to-leaf path; and \textbf{(iii)} aggregate them into the final question before answering it based on the dialogue content.

% This design elevates multi-hop reasoning from basic pointer-following to compositional reasoning: structurally, the tree mirrors a \emph{trie} where each successive edge narrows the semantic scope---from person to event to question detail---but operates in \emph{semantic space} rather than on discrete symbols, requiring meaning-level disambiguation at every branching point.

\subsection{\naen{Long-Context Multi-turn Dialogue Reinforcement Learning}}
\label{sec:rl}

We adopt GRPO~\cite{shao2024deepseekmath} with curriculum RL training and a combined reward function.

% \textbf{Group Relative Policy Optimization (GRPO).}
% \label{sec:grpo}

% Following \citep{deepseekr1, kimi1.5}, we use a combined reward function to assign $r_i$ and mitigate reward hacking.

% \textbf{Exact Match (EM) Reward.}
% Long-context dialogue QA requires precise answer extraction, so we use an Exact Match score as the trajectory reward. This provides a clear, unambiguous signal that directly optimizes for answer correctness, while avoiding the complexity of LLM-based evaluators. 
% Following \cite{deepseekr1,wang2026loongrl}, we provide a system prompt (Appendix~\ref{app:prompts}) that instructs the model to reason step-by-step and place the final answer inside \verb|\boxed{...}| for reward evaluation.
% We compute the reward as a binary exact match score between the predicted answer $\hat{a}$ and ground truth $a$, where $r_i = 1$ if $\hat{a} = a$ and $r_i = 0$ otherwise.

\textbf{Curriculum RL Training.}
% The task complexity of directly training on high-depth state trees exceeds what the model can reliably learn from scratch, leading to training instability and poor convergence. We therefore organize training into four progressive stages that gradually increase task difficulty, allowing the model to build competency incrementally. The first three stages train on the \emph{Basic StateTree} with increasing depth; the final stage transitions to the \emph{Compositional StateTree} to add compositional reasoning.
Training directly on high-depth trees is difficult for LLMs and often leads to instability. We therefore adopt a \naen{four}-stage curriculum that gradually increases task difficulty.
\begin{itemize}[nosep,leftmargin=11pt,topsep=0pt]
\item \textbf{Warmup: Direct Dialogue QA.}
We first warm up the model with RL on the original dialogue QA data, establishing basic long-context comprehension before introducing tree-structured tasks.
% \item \textbf{Stage~1--2: Basic StateTree ($D{=}2 \to 3$).}
% We train on the Basic StateTree at progressively greater depths $D{=}2$ and $D{=}3$, with each additional level introducing one more hop.
\item \naen{\textbf{Stage~1: Basic StateTree ($D{=}2$).}
We introduce the Basic StateTree at depth $D{=}2$ (4 leaves, 6 edge records scattered across sessions), requiring the model to perform cross-session retrieval and temporal discrimination as fundamental skills to reach the target leaf.}
\item \naen{\textbf{Stage~2: Basic StateTree ($D{=}3$).}
We increase the depth to $D{=}3$ (8 leaves, 14 edge records), adding an additional hop that forces the model to strengthen multi-hop reasoning.}
\item \textbf{Stage~3: Compositional StateTree ($D{=}3$).}
Stage~3 replaces the Basic StateTree with the Compositional StateTree, adding the compositional skill of accumulating step fragments along the correct path and aggregating them into the final question before answering.
\end{itemize}

\textbf{Combined Reward.}
\naen{Training with a single reward source is suboptimal: exact-match (EM) reward is overly rigid and rejects semantically correct answers that differ in surface form, while LLM-as-a-Judge suffers from scoring inconsistency and higher computational overhead (see ablation in Table~\ref{tab:reward_ablation}).}
\naen{To provide a stable and semantically faithful optimization signal,} we assign the reward $r_i$ via a combined function that integrates rule-based verification with LLM-as-a-Judge\naen{~\cite{wan2025qwenlong}}. 
% Following the system prompt (Appendix~\ref{app:prompts}), the model places its final answer $y_{\text{ans}}$ inside \verb|\boxed{...}|. 
We explicitly require the model to output its final
answer within \verb|\boxed{...}| in the training prompt (Appendix~\ref{app:prompts}), ensuring the extraction of an unambiguous answer $y_{\text{ans}}$.
%We extract $y_{\text{ans}}$ from this block and then compute:
\naen{Formally, given question $x$, extracted answer $y_{\text{ans}}$, and ground-truth answer $y_{\text{gold}}$, the reward is defined as:}
% \vspace{-2px}
\begin{equation}
r_{\phi}(x, y) = \max\big(r_{\text{EM}}(y_{\text{ans}} = y_{\text{gold}}),\; r_{\text{LLM}}(x, y_{\text{ans}}, y_{\text{gold}})\big)
% \vspace{-2px}
\end{equation}
where $r_{\text{EM}}(\cdot)$ is the indicator function enforcing exact string matching, and \naen{$r_{\text{LLM}}(\cdot) \in \{0, 1\}$ is a binary semantic equivalence score produced by a Qwen2.5-1.5B-Instruct~\citep{qwen2}. The judge operates at temperature $0$ with prompt templates (Appendix~\ref{app:locomo_eval_prompt}) to guarantee deterministic outputs.}
% The $\max$ operator serves as a logical \textit{OR} over correctness criteria, which can mitigate false negatives from rigid exact matching.}

% \textbf{Fine-Grained F1 Reward.}
% Long-context dialogue QA is inherently open-ended, so we use an F1 score as the trajectory reward. This provides dense, fine-grained feedback without the overhead of LLM-based evaluators or the brittleness of exact-match rewards, while mitigating reward hacking (e.g., the model hedges by emitting multiple candidate answers that overlap with distractors). 
% Following \cite{deepseekr1,wang2026loongrl}, we provide a system prompt (Appendix~\ref{app:prompts}) that instructs the model to reason step-by-step and place the final answer inside \verb|\boxed{...}| for reward evaluation.
% We compute the reward as the token-level F1 score between the predicted answer $\hat{a}$ and ground truth $a$.

%% file: 4_experiments.tex
\subsection{Experimental Setup}

% \textbf{Implementation Details.}
% Our curriculum comprises three stages of increasing complexity: Stage~I uses a Basic State Tree with depth $D{=}2$ (4 leaves, 6 edge records), Stage~II increases to $D{=}3$ (8 leaves, 14 edge records), and Stage~III replaces it with a Semantic Decomposition Tree at $D{=}3$, where edges carry step-level semantic fragments instead of bare pointers. All edge records are inserted within speakers' utterances at natural dialogue boundaries. Full data construction  parameters are provided in Appendix~\ref{app:data_params}.

\textbf{Training Setup.}  We run experiments on Qwen2.5-7B-Instruct~\cite{qwen2}, Qwen2.5-14B-Instruct~\cite{qwen2}, and Qwen3-8B~\cite{qwen3}. We use GRPO with group size 8, prompt batch size 64 for 7B and 8B models and 32 for 14B model, learning rate $1\times10^{-6}$, gradient clipping 1.0, and KL penalty $\beta{=}0.001$. Rollouts are sampled with temperature 0.6 and top-p$=$0.95, with a maximum output length of 4{,}096 tokens. The curriculum consists of four stages: warm-up (40 steps), Stage~1 (100 steps), Stage~2 (100 steps), and Stage~3 (60 steps). To prevent catastrophic forgetting of general reasoning capabilities, we mix 2,500 samples from the DAPO-Math dataset~\cite{yu2025dapo} into Stage~3.
The warm-up stage uses the 616 training samples in the original direct QA format without trees; Stage~1--3 augment the same 616 samples with \lib. The 770 test samples and 154 validation samples both retain the original direct QA format without any \lib augmentation.
All models are trained on 32$\times$H20 GPUs. % Detailed parameters are provided in Appendix~\ref{app:data_params}.

\textbf{Evaluation Benchmarks.} We evaluate models across two dimensions. 
(i) Long-term conversation reasoning. \naen{We use LoCoMo~\cite{maharana2024evaluating} as the in-domain (ID) benchmark, and LongMemEval~\cite{wu2025longmemeval} and PersonaMem~\cite{jiang2025know} as out-of-distribution (OOD) benchmarks to test generalization across unseen domains and extended contexts (10K--128K tokens). We follow official protocols: LoCoMo reports LLM-judged accuracy (evaluated by GPT-4o), token-level F1, and BLEU-1; LongMemEval reports LLM-judged accuracy (evaluated by GPT-4o); PersonaMem reports exact-match accuracy.}
(ii) General short-context reasoning. \naen{To verify that our long-context RL training does not induce catastrophic forgetting or degrade foundational reasoning, we additionally evaluate on standard short-context benchmarks: MMLU~\citep{hendrycks2020mmlu}, MATH-500~\citep{math-500}, and IFEval~\citep{zhou2023instruction}.}
\naen{All evaluations share a unified inference setup, i.e., temperature 0.6,} with up to 128K input tokens and 4096 output tokens.
% \textbf{(iii) Long-context retrieval}: to measure the impact of long-context RL on retrieval abilities, we evaluate on  Needle in a Haystack~\citep{needlehaystack} and RULER~\citep{ruler}.

\textbf{Baselines.} We compare our \lib-trained models against: 
\textbf{(i)} leading frontier models and long-context reasoning models, including GPT-4o\naen{~\cite{hurst2024gpt}}, QwenLong-L1-32B\naen{~\cite{wan2025qwenlong}}, and R1-Distill-Qwen-32B\naen{~\cite{guo2025deepseek}}; 
\textbf{(ii)} memory-augmented models that enhance long-context reasoning, including the SFT-based method SEALONG~\cite{li2024large} and RL-based methods LoongRL~\cite{wang2026loongrl} and RL-MemAgent~\cite{yu2026memagent}.

\vspace{-1px}
\subsection{Main Results}
\vspace{-1px}

\input{tab/locomo}

\input{tab/ood2}

\textbf{\naen{StateTree improves long-term dialogue reasoning with high data efficiency.}} 
% Table~\ref{tab:locomo_comparison} compares \lib against state-of-the-art models. The observations are as follows:
% (i) \lib improves consistently across model scales. \lib-7B improves over the base model across all metrics and even surpasses the much larger QwenLong-L1-32B. \lib-14B attains 60.91\% accuracy, surpassing GPT-4o by 8.18 and QwenLong-L1-32B by 14.55.
% (ii) \lib surpasses both SFT and RL baselines. Both \lib-7B and \lib-14B outperform the SFT-based SEALONG. Compared with RL-based LoongRL and RL-MemAgent, \lib-7B and \lib-14B lead across all three metrics.
% (iii) Temporal reasoning shows the most pronounced improvement. 
% \lib-7B and \lib-14B improve temporal F1 by +21.30 and +26.19 respectively. This suggests that state tree traversal requires comparing session timestamps at each fork.
\naen{As shown in Table~\ref{tab:locomo_comparison}, StateTree delivers consistent performance gains over both SFT and standard RL baselines across parameter scales and base architectures on LoCoMo. StateTree-14B achieves 60.91\% average accuracy, yielding a +11.82\% absolute improvement over its base model (Qwen2.5-14B-Instruct), while maintaining stable relative gains at the 7B scale. Notably, despite training on only 616 examples, it matches or exceeds the absolute performance of substantially larger models (e.g., surpassing GPT-4o by 8.18\% and QwenLong-L1-32B by 14.55\%), underscoring its high sample efficiency. The most pronounced lift occurs in Temporal reasoning (F1 +26.19 for 14B), which directly stems from the tree structure's explicit timestamp comparison at each branching step.}

% \textbf{\lib generalizes to much longer unseen long-term conversation benchmarks.}
% Since \lib is trained on the LoCoMo dataset, we evaluate OOD performance on LongMemEval~\citep{wu2025longmemeval} (Table~\ref{tab:acc_results}) and PersonaMem~\citep{jiang2025know} (Table~\ref{tab:model_comparison_3lines}).
% As shown in Table~\ref{tab:acc_results}, \lib-14B achieves 59.00\% average accuracy on LongMemEval, outperforming its base model by +13.80\% and QwenLong-L1-32B by +8.00\%. 
% On PersonaMem (Table~\ref{tab:model_comparison_3lines}), \lib-14B achieves 64.52\% average accuracy (+5.61\% over base). \lib excels at recalling update reasons (86.87\%), aligning with state tree's explicit tracking of state transitions. Both benchmarks show that \lib maintains strong performance on single-session tasks while improving multi-session reasoning.

\input{tab/length_generalization}

% \textbf{\lib generalizes to much longer unseen long-term dialogue benchmarks.}
\textbf{\naen{Robust out-of-distribution and length generalization.}}
To evaluate out-of-distribution generalization, we evaluate \lib on LongMemEval~\citep{wu2025longmemeval} and PersonaMem-128k~\citep{jiang2025know}.
% (Table~\ref{tab:acc_results})  (Table~\ref{tab:personamem_128k}).
\naen{Trained exclusively on 10K-token contexts, StateTree generalizes to unseen benchmarks with up to 128K tokens.} As shown in Table~\ref{tab:ood_results}, \lib-14B achieves 59.00\% on LongMemEval and 52.59\% on PersonaMem-128k, outperforming both base models and strong baselines including QwenLong-L1-32B and RL-MemAgent-14B. 
% LongMemEval sees the largest absolute gains (+23.60 for 7B, +13.80 for 14B), likely because its tasks demand the most explicit cross-session retrieval and temporal reasoning, which StateTree directly targets through structured multi-session state tracking. 
% \lib excels at tracking preference evolution (70.09\%) and revisiting update reasons (77.32\%), reflecting StateTree's explicit cross-session state tracking.
\naen{The widening performance gap at longer contexts (Table~\ref{tab:length_generalization}) demonstrates that the learned path-tracing strategy extrapolates effectively beyond the 10K training horizon, validating our curriculum design as an alternative to prohibitive full-length RL costs ($\mathbb{C}3$).}

\input{tab/short}

\textbf{\naen{StateTree preserves short-context abilities without capability trade-offs.}}
A common concern with long-context RL is the degradation of general reasoning. 
% Table~\ref{tab:reasoning_results_wrap} evaluates \lib on short-context reasoning and general tasks.
% \lib-7B drops only 0.2 points from the base model, a smaller degradation than competing methods. At 14B, \lib retains 80.7 against the base 81.3, matching the best RL baselines.
% Across both scales, \lib preserves base-model capabilities with minimal degradation, suggesting that our curriculum training preserves the base models' capabilities.
\naen{As shown in Table~\ref{tab:reasoning_results_wrap}, StateTree maintains competitive performance across standard short-context benchmarks (MMLU, MATH, IFEval). Notably, the 8B variant exhibits zero degradation, achieving a marginal +0.1\% average gain over the base Qwen3-8B, while the 7B and 14B models incur minimal drops of only -0.2\% and -0.6\%, respectively.} % This suggests that our curriculum training preserves the base models' capabilities.

\vspace{-2px}
\subsection{\naen{Analysis}}

% \textbf{Multi-stage curriculum RL training sustains improvements.} As shown in Figure~\ref{fig:f1_score}, average response length steadily increases throughout training, indicating that the model learns to produce detailed reasoning traces as task complexity increases. The EM reward score rises monotonically across stages with no sign of reward hacking, confirming that the exact-match reward provides a stable training signal and that the multi-stage curriculum is effective.

\begin{figure}[!t]
    \centering
    % \vspace{-1mm}
    \includegraphics[width=\textwidth]{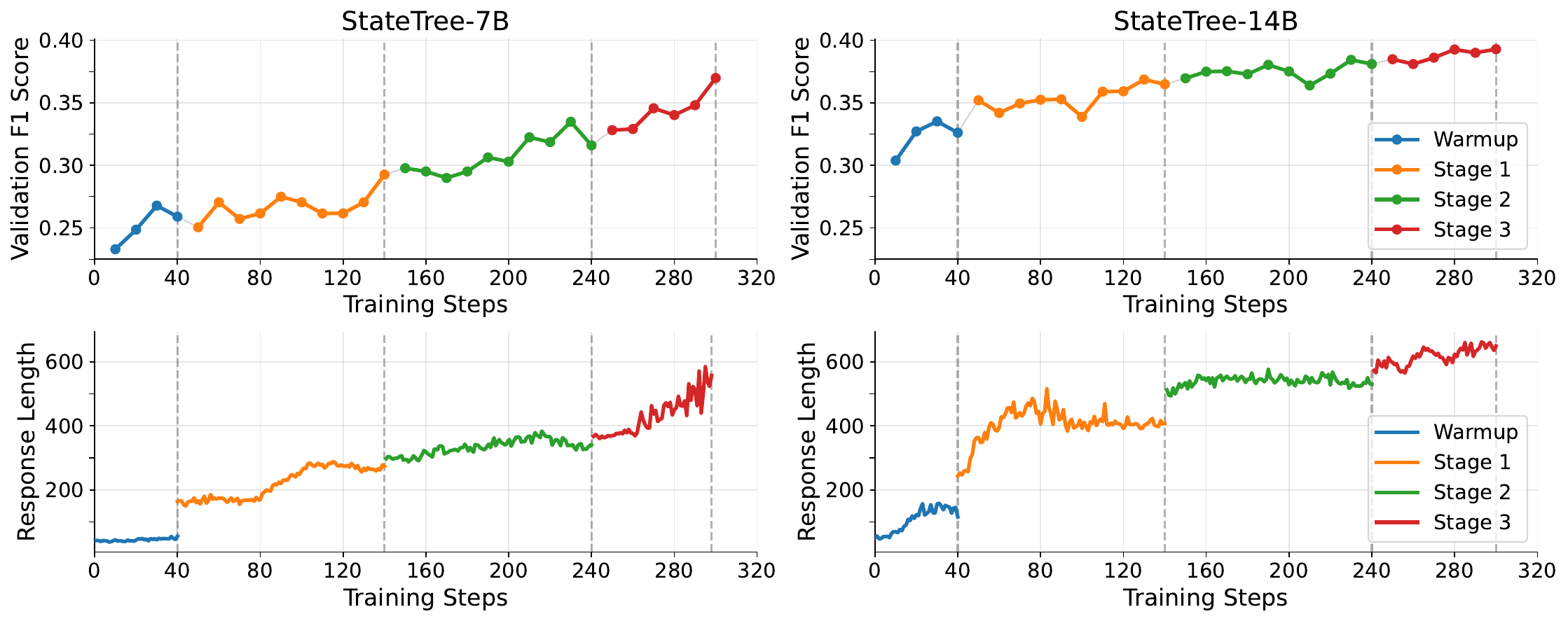}
    % \vspace{-1mm}
    \vspace{-10px}
    \caption{Reasoning F1 score and response lengths throughout RL training on the validation set.}
    \label{fig:f1_score}
    \vspace{-10px}
    % \vspace{-5mm}
\end{figure}

\textbf{Multi-stage curriculum RL training sustains improvements.} As shown in Figure~\ref{fig:f1_score}, the average F1 score on the validation set rises monotonically across training stages, and the average response length steadily increases alongside it, indicating that the model learns to produce detailed reasoning traces as task complexity increases. The sustained growth in both metrics with no sign of reward hacking confirms that the combined reward provides a stable training signal and that the multi-stage curriculum effectively scales model reasoning.

\textbf{\naen{Emergent reasoning behaviors align with task design.}} % Beyond the quantitative gains, we find that RL training with \lib enables models to develop reasoning behaviors that long-term dialogue reasoning needs. 
\naen{Qualitative analysis (Figure~\ref{fig:examples} in Section~\ref{introduction} and Appendix~\ref{app:qualitative}) reveals four emergent behaviors that directly explain the quantitative gains:}
(i) \emph{Cross-session retrieval.} Rather than stopping at the first matching session, \lib enumerates all relevant records across sessions before aggregating a complete answer.
(ii) \emph{Multi-hop reasoning.} Rather than identifying isolated facts, \lib chains multiple retrieval steps and synthesizes scattered evidence into a coherent inference.
(iii) \emph{Temporal reasoning.} Rather than relying on superficial cues such as ``recently,'' \lib extracts concrete session timestamps and performs explicit chronological comparison at each fork.
(iv) \emph{Knowledge update.} Rather than conflating old and new mentions, \lib correctly tracks the latest state by privileging information from more recent sessions.
\naen{These behaviors arise from the StateTree pseudo-task design, confirming that our auxiliary training signal successfully imparts targeted reasoning patterns that can transfer beyond the training distribution.}
% which embeds challenging reasoning patterns within a 10K training context, and generalize to 128K at inference.
%  (see Appendix~\ref{app:qualitative} for qualitative examples).

% We provided examples of each pattern in Appendix~\ref{app:qualitative}.

% \naen{\textbf{Case study.}
% Figure~\ref{fig:examples} in Section~\ref{introduction} compares the reasoning trajectories of QwenLong-L1-32B and our \lib-14B on a 128K long-term dialogue task from LongMemEval.
% When asked to count rare items across three sessions spanning two days, the 32B baseline gets lost in the multi-turn dialogue~\cite{liu2024lost,laban2026llms}, misses the 12 rare figurines mentioned in the earliest session and yields an answer of 87.
% In contrast, \lib-14B systematically retrieves all relevant mentions across sessions (12 figurines, 57 records, 5 books, 25 coins) and correctly aggregates them to 99.}
% This behavior exemplifies cross-session retrieval without positional bias, confirming that the synthetic StateTree task imparts transferable reasoning skills that enable a smaller model to outperform a much larger long-context baseline.}

% This demonstrates that synthetic StateTree training enables smaller models to outperform larger long-context baselines on cross-session reasoning.

\vspace{-2px}
\subsection{Ablation Study}
\label{sec:ablation}

\begin{wraptable}{r}{0.6\textwidth}
    \centering
    \vspace{-12pt}
    \caption{Curriculum stage ablation.}
    \label{tab:curriculum_ablation}
    \resizebox{\linewidth}{!}{
    \setlength{\tabcolsep}{2pt}
    \begin{tabular}{lccc}
        \toprule
        \textbf{Variant} & \textbf{LoCoMo} & \textbf{LongMemEval} & \textbf{PersonaMem} \\
        \midrule
        \rowcolor{gray!10}\textbf{\lib-7B (full)} & \textbf{49.61}\phantom{\scriptsize{-0.00}} & \textbf{47.40}\phantom{\scriptsize{-0.00}} & \textbf{47.30}\phantom{\scriptsize{-0.00}} \\
        % \midrule
        \quad w/o warm-up & 45.19\textcolor{red}{\scriptsize{-4.42}} & 43.00\textcolor{red}{\scriptsize{-4.40}} & 44.00\textcolor{red}{\scriptsize{-3.30}} \\
        \quad w/o Stage~1 ($D{=}2$) & 44.68\textcolor{red}{\scriptsize{-4.93}} & 42.60\textcolor{red}{\scriptsize{-4.80}} & 43.78\textcolor{red}{\scriptsize{-3.52}} \\
        \quad w/o Stage~2 ($D{=}3$) & 46.23\textcolor{red}{\scriptsize{-3.38}} & 44.20\textcolor{red}{\scriptsize{-3.20}} & 45.10\textcolor{red}{\scriptsize{-2.20}} \\
        \quad w/o Stage~3 (Compositional) & 46.75\textcolor{red}{\scriptsize{-2.86}} & 44.80\textcolor{red}{\scriptsize{-2.60}} & 45.51\textcolor{red}{\scriptsize{-1.79}} \\
        \bottomrule
    \end{tabular}
    }
    \vspace{-8pt}
\end{wraptable}
\textbf{Every curriculum stage contributes.}
Table~\ref{tab:curriculum_ablation} ablates each stage individually, where the omitted stage is skipped and the remaining stages retain their original step allocations.
Removing Stage~1 ($D{=}2$) causes the largest drop (-4.93), confirming that shallow tree navigation is the most critical foundational skill.
Removing warm-up follows ($-4.42$), showing that grounding in basic conversation QA is essential before tree tasks.
Removing Stage~2 ($D{=}3$) and Stage~3 (Compositional) degrade performance by -3.38 and -2.86, respectively, indicating their roles in strengthening multi-hop temporal reasoning and enabling compositional reasoning.
% Overall, the full curriculum lifts the base model by 5.32 accuracy points on LoCoMo, with commensurate gains on OOD benchmarks.
Appendix~\ref{app:category_ablation} provides a per-category breakdown that further corroborates the curriculum design.
\naen{This sustained uplift validates our staged design: by progressively introducing foundational comprehension, temporal discrimination, and compositional reasoning, the curriculum enables stable skill accumulation without catastrophic forgetting, effectively distilling transferable long-context primitives from a compact training set.}

\begin{wraptable}{r}{0.55\textwidth}
    \centering
    \vspace{-10pt}
    \caption{Temporal discrimination ablation.}
    \label{tab:locomo_ablation}
    \resizebox{\linewidth}{!}{
    \setlength{\tabcolsep}{3pt}
    \begin{tabular}{lccc}
        \toprule
        \textbf{Variant} & \textbf{LoCoMo} & \textbf{LongMemEval} & \textbf{PersonaMem} \\
        \midrule
        \rowcolor{gray!10}Qwen2.5-7B-Instruct & 44.29\phantom{\scriptsize{-0.00}} & 23.80\phantom{\scriptsize{-0.000}} & 40.48\phantom{\scriptsize{-0.00}} \\% 341/770, 119/500, 1104/2727
        \quad w/ CoT prompt & 45.19\textcolor{lightgreenzh}{\scriptsize{+0.90}} & 24.60\textcolor{lightgreenzh}{\scriptsize{+0.80}}\phantom{\scriptsize{0}} & 41.11\textcolor{lightgreenzh}{\scriptsize{+0.63}} \\% 348/770, 123/500, 1121/2727
        \midrule
        \rowcolor{gray!10}\textbf{\lib-7B} & \textbf{49.61}\phantom{\scriptsize{-0.00}} & \textbf{47.40}\phantom{\scriptsize{-0.000}} & \textbf{47.30}\phantom{\scriptsize{-0.00}} \\% 382/770, 237/500, 1290/2727
        \quad w/ \naen{Independent Random Keys} & 45.45\textcolor{red}{\scriptsize{-4.16}} & 25.20\textcolor{red}{\scriptsize{-22.20}} & 41.03\textcolor{red}{\scriptsize{-6.27}} \\% 350/770, 126/500, 1119/2727
        \quad  w/ Single Session & 47.79\textcolor{red}{\scriptsize{-1.82}} & 42.80\textcolor{red}{\scriptsize{-4.60}}\phantom{\scriptsize{0}} & 43.96\textcolor{red}{\scriptsize{-3.34}} \\% 368/770, 214/500, 1199/2727
        \quad  w/ Entity Keys & 48.44\textcolor{red}{\scriptsize{-1.17}} & 45.60\textcolor{red}{\scriptsize{-1.80}}\phantom{\scriptsize{0}} & 45.98\textcolor{red}{\scriptsize{-1.32}} \\% 373/770, 228/500, 1254/2727
        
        \bottomrule
    \end{tabular}
    }
    \vspace{-10pt}
\end{wraptable}
\textbf{Task structure drives the training signal.}
% We design four variants to isolate the contribution of each structural design choice (Table~\ref{tab:locomo_ablation}).
\naen{To confirm that StateTree's efficacy stems from the synergistic integration of key-paired forks, cross-session scattering, and identifier opacity---rather than from merely inserting structured records or prompting longer reasoning, we design} four variants to isolate the contribution of each structural design choice (Table~\ref{tab:locomo_ablation}).
\begin{itemize}[nosep,leftmargin=11pt,topsep=0pt]
    \item \emph{w/ CoT prompt}: \naen{While Figure~\ref{fig:f1_score} shows that StateTree training naturally elicits progressively longer reasoning traces, prompting the base model via a modified system prompt (Appendix~\ref{app:cot_prompt}) that instructs it to ``think step by step'' yields only marginal gains. This confirms that generic CoT cannot substitute for explicit structural navigation. The core bottleneck lies in unstructured cross-session retrieval and temporal discrimination, not insufficient reasoning depth.}% prompting the base model to ``think step by step'' yields only marginal gains (+0.90 on LoCoMo), confirming that longer generic reasoning chains do not address the core challenge of cross-session retrieval and temporal discrimination.

    \item \naen{\emph{w/ Independent Random Keys}: assigning independent random keys to each fork's competing edge records eliminates shared identifiers,} preventing the model from locating competing records or performing temporal comparison. \naen{The resulting performance demonstrates that StateTree's gains originate specifically from the shared-key mechanism enabling explicit fork resolution, rather than from merely injecting records or arbitrary syntactic noise during RL training.} % Gains remain barely above the CoT-prompted base model, confirming key pairing as a prerequisite for a useful training signal.
    \item \emph{w/ Single Session}: preserving key pairing but placing all records within one session eliminates cross-session retrieval. \naen{The intermediate performance confirms that cross-session distribution and temporal discrimination provide complementary, non-redundant learning signals.}% This variant outperforms \emph{w/o Key Pairing} but falls below \emph{w/ Entity Keys}, showing that cross-session distribution provides an independent training signal.
    \item \emph{w/ Entity Keys}: replacing random UUIDs with common words from a word pool (Appendix~\ref{app:entity_keys}) captures 80\% of the full gain. \naen{The remaining deficit reveals that semantically interpretable keys invite shallow lexical shortcuts, whereas opaque identifiers force genuine temporal reasoning.}% The remaining gap (+1.17 on LoCoMo) indicates that opaque keys are important---semantically interpretable keys allow shallow lexical associations, weakening temporal discrimination.
\end{itemize}

% Together, these results confirm that the training signal stems from the full structural design (key pairing, cross-session distribution, and opaque identifiers) rather than from merely inserting structured records or prompting longer reasoning.

% \begin{wraptable}{r}{0.48\textwidth}
%     \centering
%     \vspace{-15pt}
%     \caption{Effect of tree depth.}
%     \label{tab:depth_ablation}
%     \resizebox{\linewidth}{!}{
%     \setlength{\tabcolsep}{3pt}
%     \begin{tabular}{lccccc}
%         \toprule
%         \textbf{Depth} & \textbf{Leaves} & \textbf{Edges} & \textbf{LoCoMo} & \textbf{LongMemEval} & \textbf{PersonaMem} \\
%         \midrule
%         $D{=}1$ & 2 & 2 & 51.95 & 48.40 & 57.72 \\
%         $D{=}2$ & 4 & 6 & 56.49 & 54.20 & 61.29 \\
%         $D{=}3$ (default) & 8 & 14 & \textbf{60.91} & \textbf{59.00} & \textbf{64.52} \\
%         $D{=}4$ & 16 & 30 & 59.74 & 57.60 & 63.67 \\
%         \bottomrule
%     \end{tabular}
%     }
%     \vspace{-10pt}
% \end{wraptable}
% \textbf{Tree depth controls task difficulty.}
% Table~\ref{tab:depth_ablation} varies the final-stage tree depth.
% $D{=}1$ provides insufficient challenge with only a single fork. $D{=}2$ and $D{=}3$ yield progressively larger gains across all benchmarks. $D{=}3$ achieves the best results.
% At $D{=}4$, performance slightly degrades: 30 edge records across $\sim$13 sessions increase record density, making temporal discrimination noisier.
% This suggests tree depth should be calibrated to context length; $D{=}3$ is optimal for 16K contexts.

\begin{wraptable}{r}{0.55\textwidth}
    \centering
    \vspace{-12pt}
    \caption{Reward function ablation.}
    \vspace{-5pt}
    \label{tab:reward_ablation}
    \resizebox{\linewidth}{!}{
    \setlength{\tabcolsep}{3pt}
    \begin{tabular}{lccc}
        \toprule
        \textbf{Reward} & \textbf{LoCoMo} & \textbf{LongMemEval} & \textbf{PersonaMem} \\
        \midrule
        \rowcolor{gray!10}\textbf{Combined Reward} & \textbf{49.61}\phantom{\scriptsize{-0.00}} & \textbf{47.40}\phantom{\scriptsize{-0.00}} & \textbf{47.30}\phantom{\scriptsize{-0.00}} \\% 382/770, 237/500, 1290/2727
        
        Exact Match only & 47.79\textcolor{red}{\scriptsize{-1.82}} & 46.80\textcolor{red}{\scriptsize{-0.60}} & 46.68\textcolor{red}{\scriptsize{-0.62}} \\% 368/770, 234/500, 1273/2727
        LLM-as-a-Judge only & 48.70\textcolor{red}{\scriptsize{-0.91}} & 46.60\textcolor{red}{\scriptsize{-0.80}} & 46.75\textcolor{red}{\scriptsize{-0.55}} \\% 375/770, 233/500, 1275/2727
        Token-level F1 & 47.14\textcolor{red}{\scriptsize{-2.47}} & 45.00\textcolor{red}{\scriptsize{-2.40}} & 45.29\textcolor{red}{\scriptsize{-2.01}} \\% 363/770, 225/500, 1235/2727
        Two-way Substr.\ EM & 44.68\textcolor{red}{\scriptsize{-4.93}} & 42.40\textcolor{red}{\scriptsize{-5.00}} & 43.31\textcolor{red}{\scriptsize{-3.99}} \\% 344/770, 212/500, 1181/2727
        ROUGE-L & 42.60\textcolor{red}{\scriptsize{-7.01}} & 39.40\textcolor{red}{\scriptsize{-8.00}} & 41.29\textcolor{red}{\scriptsize{-6.01}} \\% 328/770, 197/500, 1126/2727
        
        \bottomrule
    \end{tabular}
    }
    \vspace{-15pt}
\end{wraptable}
\textbf{Combined reward is critical for \naen{stable optimization}.}
% Table~\ref{tab:reward_ablation} compares reward functions under the same curriculum and training configuration.
% Our combined reward integrates exact-match verification with an LLM-as-a-Judge evaluator via maximum selection, directly targeting answer correctness while mitigating false negatives from strict string matching.
% Exact Match alone is overly rigid, rejecting semantically correct answers that differ in surface form.
% LLM-as-a-Judge alone is computationally expensive yet underperforms our combined approach, as standalone LLM judgments are inconsistent.
% Token-level F1 and Two-way Substring EM provide denser feedback but reward partial matches that may be wrong.
% ROUGE-L performs worst: as a purely lexical metric, it rewards superficial overlap regardless of correctness.
\naen{Table~\ref{tab:reward_ablation} compares reward designs under identical training settings. The combined reward balances strict correctness with semantic flexibility. EM alone rejects valid paraphrases due to surface-form rigidity, while standalone LLM-Judge suffers from scoring inconsistency and higher compute overhead. Dense lexical metrics (Token F1, Substring EM) and ROUGE-L underperform by rewarding incorrect partial matches or superficial n-gram overlap. Thus, the max-selection mechanism is essential to provide a dense yet reliable training signal aligned with answer correctness.}

%% file: tab/locomo.tex
\begin{table}[t]
    \centering
    \caption{Main results on the LoCoMo benchmark across different task dimensions.}
    \label{tab:locomo_comparison}
    \resizebox{\textwidth}{!}{
    \setlength{\tabcolsep}{3pt}
    \begin{tabular}{lcccccccccccc>{\cellcolor{gray!10}}c>{\cellcolor{gray!10}}c>{\cellcolor{gray!10}}c}
        \toprule
        \multirow{2}{*}{\bf Model} & \multicolumn{3}{c}{\textbf{Multi Hop}} & \multicolumn{3}{c}{\textbf{Temporal}} & \multicolumn{3}{c}{\textbf{Open Domain}} & \multicolumn{3}{c}{\textbf{Single Hop}} & \multicolumn{3}{c}{\textbf{Average}} \\
        \cmidrule(lr){2-4} \cmidrule(lr){5-7} \cmidrule(lr){8-10} \cmidrule(lr){11-13} \cmidrule(lr){14-16}
        & \textbf{ACC} & \textbf{F1} & \textbf{BLEU} & \textbf{ACC} & \textbf{F1} & \textbf{BLEU} & \textbf{ACC} & \textbf{F1} & \textbf{BLEU} & \textbf{ACC} & \textbf{F1} & \textbf{BLEU} & \textbf{ACC} & \textbf{F1} & \textbf{BLEU} \\
        \midrule
        GPT-4o    & 68.84 & \bf 32.25 & \bf 23.10 & 14.47 & 10.25 & 10.45 & 30.00 & 18.47 & 16.80 & \bf 64.54 & \bf 49.34 & \bf 40.19 & \bf 52.73 & \bf 36.20 & \bf 29.47  \\
        QwenLong-L1-32B &64.49 & 29.53 & 22.27 & 34.59 & \bf 24.35 & \bf 20.19 & 48.00 & \bf 29.43 & \bf 27.02 & 44.68 & 30.37 & 26.20 & 46.36 & 28.92 & 24.31 \\
        R1-Distill-Qwen-32B & \bf 70.29 & 26.80 & 17.45 & \bf 33.33 & 15.98 & 12.35 & \bf 52.00 & 25.19 & 20.98 & 52.01 & 23.04 & 18.00 & 51.43 & 22.40 & 16.93\\
        % R1-Distill-LLaMa-70B & 63.77 & 29.00 & 18.91 & \bf 33.33 & 19.11 & 15.25 & \bf  56.00 & 25.73 & 20.76 & 49.16 &26.43 & 21.58 & 48.95& 25.32 & 19.72\\
        \midrule  
        Qwen2.5-7B-Instruct  & 63.77  & 21.34 & 15.23  & 23.90  & 12.67 & 10.58  & 50.00  & 16.88 & 14.08  & 44.92  & 21.96 & 17.72   & 44.29  & 19.60 & 15.56  \\
        SEALONG-7B & 63.77 & 27.50 & 20.65 & 20.75 & 15.16 & 11.72 & 48.00 & 25.80 & 23.53 & 47.99 & 32.01 & 28.38 & 45.19 & 27.32 & 23.24  \\
        LoongRL-7B & 60.87 & 28.84 & 19.53 & 25.16 & 18.25 & 15.08 & \bf 56.00 & 23.56 & 19.68 & 47.99 & 32.92 & 28.41 & 46.10 & 28.55 & 23.50  \\
        RL-MemAgent-7B & \bf 64.49 & 30.62 & 21.88 & 31.45 & 16.36 & 12.31 & 46.00 & 26.26 & 22.41 & 49.14 & 32.75 & 27.07 & 48.03  & 28.56  & 22.79 \\
        \textbf{\lib-7B}  & \bf 64.49 & \bf 31.20 & \bf 25.38 & \bf 32.70 & \bf 33.97 & \bf 29.36 & \bf 56.00 & \bf 36.11 & \bf 32.31 & \bf 50.35 & \bf 36.00 & \bf 32.18 & \bf 49.61  & \bf 34.73  & \bf 30.39 \\
        
        \midrule    
        Qwen2.5-14B-Instruct & 67.39 & 31.98 & 24.07 & 34.59 & 14.53 & 11.17 & 62.00 & 27.31 & 23.83 & 47.04 & 34.32 & 30.16 & 49.09 & 29.36 & 24.74 \\
        SEALONG-14B & 69.57 & 35.64 & 25.65 & 32.70 & 15.85 & 12.95 & 58.00 & 26.37 & 22.08 & 47.28 & 32.73 & 28.76 & 48.96 & 29.36 & 24.50 \\
        LoongRL-14B & 69.57 & 33.20 & 25.82 & 37.74 & 20.53 & 23.25 & 62.00 & 34.14 & 31.10 & 52.25 & 35.76 & 30.25 & 52.99 & 32.05 & 28.07 \\
        RL-MemAgent-14B & 67.39 & 31.96 & 25.01 & 37.11 & 14.55 & 11.13 & 60.00 & 33.04 & 30.10 & 50.59 & 34.62 & 29.39 & 51.43 & 29.90 & 24.88 \\
        \textbf{\lib-14B} & \bf 75.36 & \bf 38.40 & \bf 27.90 & \bf 41.51 & \bf 40.72 & \bf 33.56 & \bf 66.00 & \bf 39.65 & \bf 37.18 & \bf 62.88 & \bf 38.37 & \bf 33.26 & \bf 60.91 & \bf 38.94 & \bf 32.62 \\

        \midrule 
        Qwen3-8B & 64.49 & 31.32 & 22.13 & 29.56 & 18.91 & 15.31 & 48.00 & 21.4 & 18.55 & 43.03 & 26.37 & 23.02 & 43.64 & 25.39  & 20.98 \\
        \textbf{\lib-8B}& \bf 70.29 & \bf 36.07 & \bf 25.58 & \bf 37.74 & \bf 31.78 & \bf 29.25 & \bf 56.00 & \bf 36.36 & \bf 33.08 & \bf 51.06 & \bf 36.75 & \bf 31.89 & \bf 52.08  & \bf 35.58  & \bf 30.29  \\
        \bottomrule
    \end{tabular}}
    
    \vspace{-8px}
\end{table}

%% file: tab/ood2.tex
\begin{table*}[t]
    \centering
    \caption{OOD generalization accuracy (\%) on LongMemEval and PersonaMem-128k benchmarks.}
    \vspace{-5px}
    \label{tab:ood_results}
    \resizebox{\textwidth}{!}{
    \setlength{\tabcolsep}{2pt}
    \small
    % \dashlinedash=2pt
    % \dashlinegap=2pt
    \begin{tabular}{lcccccc>{\cellcolor{gray!10}}c:ccccccc>{\cellcolor{gray!10}}c}
        \toprule
        & \multicolumn{7}{c}{\textbf{LongMemEval (128k)}} & \multicolumn{8}{c}{\textbf{PersonaMem (128k)}} \\
        \cmidrule(lr){2-8} \cmidrule(lr){9-16}
        \textbf{Model} &
        \textbf{Temp.} &
        \makecell{\textbf{Multi-}\\\textbf{Ses.}} &
        \makecell{\textbf{Know.}\\\textbf{Upd.}} &
        \makecell{\textbf{SS-}\\\textbf{User}} &
        \makecell{\textbf{SS-}\\\textbf{Asst.}} &
        \makecell{\textbf{SS-}\\\textbf{Pref.}} &
        \textbf{Avg.} &
        \makecell{\textbf{Latest}\\\textbf{Pref.}} &
        \makecell{\textbf{New}\\\textbf{Scen.}} &
        \makecell{\textbf{Align.}\\\textbf{Rec.}} &
        \makecell{\textbf{Shared}\\\textbf{Fact}} &
        \makecell{\textbf{Revisit}\\\textbf{Reas.}} &
        \makecell{\textbf{New}\\\textbf{Idea}} &
        \makecell{\textbf{Track}\\\textbf{Evol.}} &
        \textbf{Avg.} \\

        \midrule
        QwenLong-L1-32B & \bf 47.37 & \bf 35.34 & \bf 56.41 & \bf 80.00 & \bf 67.86 & \bf 23.33 & \bf 51.00 & \bf 57.51 & \bf 48.83 & \bf 47.56 & \bf 66.08 & \bf 81.04 & \bf 25.48 & \bf 58.06 & \bf 52.40 \\
        R1-Distill-Qwen-32B & 21.80 & 13.53 & 50.00 & 34.29 & 57.14 & 10.00 & 29.00 & 33.03 & 21.60 & 33.52 & 36.84 & 63.57 & 22.39 & 66.57 & 37.62 \\

        \midrule

        Qwen2.5-7B-Instruct & 20.30 & 8.27 & 52.56 & 25.71 & 37.50 & 3.33 & 23.80 & 37.41 & 36.15 & 45.56 & 47.95 & 62.08 & 15.64 & 62.76 & 40.48 \\
        SEALONG-7B & 28.57 & 28.57 & 64.10 & 75.71 & 76.79 & 16.67 & 45.40 & 37.99 & 39.44 & 46.99 & 47.37 & 64.68 & 16.80 & 63.64 & 41.66 \\
        LoongRL-7B & 24.06 & 5.26 & 55.13 & 37.14 & 42.86 & 3.33 & 26.60 & 29.56 & 23.00 & 39.83 & 35.09 & 41.26 & 21.81 & 37.54 & 31.39 \\
        RL-MemAgent-7B & 28.57 & 17.29 & 58.97 & \bf 78.57 & 76.79 & 13.33 & 41.80 & 43.07 & 33.33 & 45.27 & 49.12 & 65.06 & 23.17 & 62.46 & 43.78 \\
        \lib-7B & \bf 29.32 & \bf 31.58 & \bf 67.95 & 75.71 & \bf 76.79 & \bf 23.33 & \bf 47.40 & \bf 46.42 & \bf 39.91 & \bf 51.00 & \bf 49.71 & \bf 68.03 & \bf 23.36 & \bf 69.21 & \bf 47.30 \\

        \midrule

        Qwen2.5-14B-Instruct & 37.59 & 27.82 & 67.95 & 65.71 & 66.07 & 10.00 & 45.20 & 43.76 & 38.03 & 52.15 & 57.31 & 68.03 & 18.34 & 64.52 & 45.40 \\
        SEALONG-14B & 40.60 & 30.08 & 67.95 & 72.86 & 66.07 & 13.33 & 47.80 & 45.84 & 38.97 & 51.29 & 61.40 & 68.77 & 15.44 & 66.28 & 46.02 \\
        LoongRL-14B & 40.60 & 36.84 & 69.23 & 81.43 & \bf 89.29 & 23.33 & 54.20 & 48.15 & 40.85 & 56.16 & 60.23 & 68.77 & 18.15 & 67.16 & 48.07 \\
        RL-MemAgent-14B & \bf 42.11 & 29.32 & 67.95 & 85.71 & 82.14 & 20.00 & 52.00 & 47.69 & 39.44 & 53.30 & 60.82 & 71.00 & 21.04 & 66.28 & 48.15 \\
        \lib-14B & \bf 42.11 & \bf 48.12 & \bf 70.51 & \bf 87.14 &  85.71 & \bf 36.67 & \bf 59.00 & \bf 51.62 & \bf 46.48 & \bf 59.89 & \bf 64.33 & \bf 77.32 & \bf 23.55 & \bf 70.09 & \bf 52.59 \\

        \midrule
        Qwen3-8B & 35.34 & 39.10 & 66.67 & 72.86 & 69.64 & 16.67 & 49.20 & 53.81 & 39.44 & 37.82 & 48.54 & 69.52 & 16.60 & 58.36 & 45.36 \\
        \textbf{\lib-8B} & \bf 41.00 & \bf 47.47 & \bf 78.21 & \bf 85.45 & \bf 83.93 & \bf 26.09 & \bf 58.66 & \bf 58.31 & \bf 46.95 & \bf 41.26 & \bf 46.78 & \bf 72.86 & \bf 21.04 & \bf 69.79 & \bf 50.31 \\

        \bottomrule
    \end{tabular}}
    \vspace{-15px}
\end{table*}

%% file: tab/length_generalization.tex
\begin{wraptable}{r}{0.58\textwidth}
    \centering
    % \vspace{-12pt}
    \caption{Average accuracy~(\%) across context lengths.}
    \vspace{-2px}
    \label{tab:length_generalization}
    % \footnotesize
    \setlength{\tabcolsep}{3pt}
    \resizebox{\linewidth}{!}{
    \begin{tabular}{llllc}
        \toprule
        \multirow{2}{*}{\bf Model} & \textbf{LoCoMo} & \multicolumn{2}{c}{\textbf{PersonaMem}} & \textbf{LongMemEval} \\
        \cmidrule(lr){2-2}\cmidrule(lr){3-4}\cmidrule(lr){5-5}
         & \multicolumn{1}{c}{\bf 10K} & \multicolumn{1}{c}{\bf 32K} & \multicolumn{1}{c}{\bf 128K} & \multicolumn{1}{c}{\bf 128K} \\
        \midrule
        Qwen2.5-7B-Instruct & 44.29 & 53.14 & 40.48 & 23.80\phantom{\scriptsize{-0.000}} \\
        \textbf{\lib-7B} & \textbf{49.61}\textcolor{lightgreenzh}{\scriptsize{+5.32}} & \textbf{61.29}\textcolor{lightgreenzh}{\scriptsize{+8.15}} & \textbf{47.30}\textcolor{lightgreenzh}{\scriptsize{+6.82}} & \textbf{47.40}\textcolor{lightgreenzh}{\scriptsize{+23.60}} \\
        \midrule
        Qwen2.5-14B-Instruct & 49.09 & 58.91 & 45.40 & 45.20\phantom{\scriptsize{-0.000}} \\
        \textbf{\lib-14B} & \textbf{60.91}\textcolor{lightgreenzh}{\scriptsize{+11.82}} & \textbf{64.52}\textcolor{lightgreenzh}{\scriptsize{+5.61}} & \textbf{52.59}\textcolor{lightgreenzh}{\scriptsize{+7.19}} & \textbf{59.00}\textcolor{lightgreenzh}{\scriptsize{+13.80}} \\
        \midrule
        Qwen3-8B & 43.64 & 58.23 & 45.36 & 49.20\phantom{\scriptsize{-0.000}} \\
        \textbf{\lib-8B} & \bf 52.08\textcolor{lightgreenzh}{\scriptsize{+8.44}} & \bf 62.82\textcolor{lightgreenzh}{\scriptsize{+4.59}} & \bf 50.31\textcolor{lightgreenzh}{\scriptsize{+4.95}} & \bf 58.66\textcolor{lightgreenzh}{\scriptsize{+9.46}}\phantom{\scriptsize{0}} \\
        \bottomrule
    \end{tabular}
    }
    % \vspace{-10pt}
\end{wraptable}

%% file: tab/short.tex
\begin{wraptable}{r}{0.5\textwidth}
\centering
\vspace{-5px}
\caption{Short-context reasoning performance (\%).}
% \vspace{-5px}

\label{tab:reasoning_results_wrap}
\resizebox{\linewidth}{!}{
\setlength{\tabcolsep}{3pt}
\begin{tabular}{lccc>{\cellcolor{gray!10}}c}
\toprule

\bf Model & \textbf{MMLU} & \textbf{MATH} & \textbf{IFEval} & \textbf{Avg.} \\ 
\midrule
% GPT-4o                     & \textbf{88.7} & 74.6 & \textbf{84.3} & 82.5 \\
% QwenLong-L1-32B             & 78.5 & \textbf{95.2} & 78.6 & 84.1 \\
% R1-Distill-Qwen-32B         & 80.5 & 94.3 & 72.5 & 82.4 \\
% R1-Distill-LLaMa-70B        & 82.4 & 94.5 & 79.3 & \textbf{85.4} \\
% \midrule
Qwen2.5-7B-Instruct         & 73.4\phantom{\scriptsize{-0.00}} & \bf 76.0\phantom{\scriptsize{-0.00}} & \textbf{71.2}\phantom{\scriptsize{-0.00}} & \bf 73.5\phantom{\scriptsize{-0.00}} \\
% SEALONG-7B                  & 73.0 & 76.5 & 69.0 & 72.8 \\
% LoongRL-7B                  & \textbf{76.2} & \textbf{78.0} & 70.9 & \textbf{75.0} \\
% RL-MemAgent-7B              & 73.0 & 76.0 & 70.0 & 73.0 \\
\textbf{\lib-7B}            & \bf 73.5\textcolor{lightgreenzh}{\scriptsize{+0.10}} & \bf 76.0\textcolor{lightgreenzh}{\scriptsize{+0.00}} & 70.5\textcolor{red}{\scriptsize{-0.70}} & 73.3\textcolor{red}{\scriptsize{-0.20}} \\
\midrule
Qwen2.5-14B-Instruct        & 79.4\phantom{\scriptsize{-0.00}} & \textbf{83.4}\phantom{\scriptsize{-0.00}} & \textbf{81.0}\phantom{\scriptsize{-0.00}} & \textbf{81.3}\phantom{\scriptsize{-0.00}} \\
% SEALONG-14B                 & 79.0 & 83.5 & 79.0 & 80.5 \\
% LoongRL-14B                 & \textbf{80.5} & 83.2 & 78.4 & 80.7 \\
% RL-MemAgent-14B             & 79.2 & 83.0 & 80.0 & 80.7 \\
\textbf{\lib-14B}           & \bf 79.8\textcolor{lightgreenzh}{\scriptsize{+0.40}} & 83.0\textcolor{red}{\scriptsize{-0.40}} & 79.3\textcolor{red}{\scriptsize{-1.70}} & 80.7\textcolor{red}{\scriptsize{-0.60}} \\

\midrule
Qwen3-8B & 76.9\phantom{\scriptsize{-0.00}} & \bf 78.2\phantom{\scriptsize{-0.00}} & 85.0\phantom{\scriptsize{-0.00}} & 80.0\phantom{\scriptsize{-0.00}} \\
\textbf{\lib-8B} & \bf 77.0\textcolor{lightgreenzh}{\scriptsize{+0.10}} & 78.1\textcolor{red}{\scriptsize{-0.10}} & \bf 85.2\textcolor{lightgreenzh}{\scriptsize{+0.20}} & \bf 80.1\textcolor{lightgreenzh}{\scriptsize{+0.10}} \\
\bottomrule
\end{tabular}
}
\vspace{-10px}
\end{wraptable}

%% file: 6_conclusion.tex
We present \lib, a data-driven RL pseudo-task that 
% embeds StateTree tasks within authentic dialogues to train cross-session retrieval, temporal reasoning, knowledge update, and compositional multi-hop reasoning. A structured curriculum with combined-reward GRPO enables stable training from only 616 examples at 10K context. 
\naen{reframes long-term dialogue reasoning as a temporally constrained path-search problem. By embedding tree-structured auxiliary tasks into authentic multi-session dialogues and optimizing them via a structured curriculum with combined-reward GRPO, StateTree elicits cross-session retrieval, temporal reasoning, knowledge update, and compositional multi-hop reasoning from only 616 examples at 10K context.}
% \lib substantially improves Qwen2.5-7B/14B-Instruct on LoCoMo, LongMemEval, and PersonaMem: \lib-14B reaches 60.91\% on LoCoMo (+11.82), surpassing QwenLong-L1-32B and R1-Distill-LLaMa-70B, while \lib-7B gains up to +23.60 on LongMemEval. 
\naen{Empirically, it yields substantial gains across both in-domain and out-of-domain benchmarks, even surpassing larger baselines.}
% Models generalize from 10K to 128K tokens, preserve short-context abilities, and exhibit emergent reasoning behaviors aligned with the StateTree design.
\naen{Crucially, the trained models generalize from 10K to 128K contexts without degrading short-context capabilities and exhibit emergent reasoning behaviors aligned with the StateTree design. StateTree establishes a scalable, data-efficient paradigm for long-term dialogue reasoning.}

%% file: 7_appendix.tex
\appendix
\newpage
% \section*{Appendix}

\section{Limitations and Broader Impacts}
\label{app:limitations}

\subsection{Limitations}

Our work has several limitations. First, all training data are derived from a single English dialogue dataset (LoCoMo), so the generalization of \lib to other languages or dialogue domains remains to be validated. Second, the Compositional StateTree relies on an LLM (GPT-4o) to decompose questions and generate distractors; while we enforce structural validation, the quality of decomposition is bounded by the LLM's capability and may introduce subtle semantic biases. Finally, we use fixed binary-tree depths ($D{=}2$ and $D{=}3$) throughout training; adaptive depths or dynamic tree structures could further improve efficiency but are not explored here. 

\subsection{Broader Impacts}

\lib aims to improve the reliability of personalized assistants in long-term interactions, which can enhance user experience in education, healthcare, and personal productivity. However, more powerful reasoning capabilities could also be misused to extract sensitive information from conversation histories or enable unauthorized surveillance. We encourage developers to deploy such systems with appropriate privacy safeguards, user consent mechanisms, and data retention policies.

\section{Licenses and Asset Usage}
\label{app:licenses}

\subsection{Existing Assets}

All existing datasets and models used in this paper are properly cited. LoCoMo~\citep{maharana2024evaluating}, LongMemEval~\citep{wu2025longmemeval}, and PersonaMem are released for research purposes. Qwen2.5-7B-Instruct, Qwen2.5-14B-Instruct, and Qwen3-8B~\citep{qwen2} are released under the Qwen License. Qwen2.5-1.5B-Instruct, used as the LLM judge, is under the same license. GPT-4o is accessed via the OpenAI API.

\subsection{LLM Usage Declaration}
\label{app:llm_usage}

LLMs are used as non-standard components in two parts of our methodology. (i)~\textbf{Data construction:} We use GPT-4o (temperature 0.7) to decompose target questions into semantic axes and generate distractor steps for the Compositional StateTree (Section~\ref{sec:data_construction_pipeline} and Appendix~\ref{app:decomp_prompt}). (ii)~\textbf{Reward evaluation:} We use Qwen2.5-1.5B-Instruct (temperature 0) as an LLM-as-a-Judge evaluator within the combined reward function (Section~\ref{sec:rl}).

\section{Dataset Construction Details}
\label{app:dataset}

\subsection{Source Data}
All training data are derived from the LoCoMo dataset~\citep{maharana2024evaluating}, which contains multi-session dialogues between two speakers. Each conversation comprises approximately 13 sessions with interleaved \texttt{DATE:} markers indicating session timestamps. We use the same 616 conversation--question pairs across all four curriculum stages; only the \emph{task formulation} changes between stages.

\subsection{Dataset Split}
\label{app:dataset_split}

We filter out adversarial and empty-answer questions, then split the remaining pairs 50/50. The first half uses an 80/20 train/validation split (616 and 154 samples), and the second half forms the held-out test set.

The validation and test sets use the original direct QA format (identical to the Warm-up stage) without any tree augmentation or record insertion. This ensures that evaluation measures the model's end-task performance on clean conversation data. The same training set of 616 samples is reused across all four curriculum stages; only the task formulation (tree structure and prompt) changes between stages.

\subsection{Dataset Statistics and Construction Parameters}

Table~\ref{tab:dataset_stats} summarizes the dataset statistics and data construction parameters for each curriculum stage.

\begin{table}[h]
    \centering
    \caption{Dataset statistics and data construction parameters for each curriculum stage. ``UUID format'' refers to the identifier format used for tree node keys. ``Insert mode'' specifies where records are placed within the dialogue text.}
    \label{tab:dataset_stats}
    \resizebox{\textwidth}{!}{
    \begin{tabular}{lccccc}
        \toprule
        \textbf{Parameter} & \textbf{Warm-up} & \textbf{Stage~1} & \textbf{Stage~2} & \textbf{Stage~3} \\
        \midrule
        Tree type           & None (direct QA)         & Basic StateTree    & Basic StateTree    & Compositional StateTree \\
        Tree depth $D$      & --           & 2                   & 3                   & 3 \\
        Leaves              & --           & 4                   & 8                   & 8 \\
        Edge records        & 0            & 6                   & 14                  & 14 \\
        Samples             & 616          & 616                 & 616                 & 616 \\
        Avg. length (tokens) & $\sim$10K  & $\sim$10K           & $\sim$10K           & $\sim$10K \\
        UUID format         & --           & Standard UUID4      & Standard UUID4      & Standard UUID4 \\
        Record format       & --           & \texttt{\{UUID: VALUE\}}  & \texttt{\{UUID: VALUE\}}  & \texttt{\{UUID: \{step, next\}\}} \\
        Insert mode         & --           & Inside speaker quotes & Inside speaker quotes & Inside speaker quotes \\
        Temporal discrimination & --       & Yes (session DATE)  & Yes (session DATE)  & Yes (session DATE) \\
        Distractor pool size & --          & 64                  & 64                  & 8 (from decomposition) \\
        \bottomrule
    \end{tabular}
    }
\end{table}

\subsection{Question Decomposition}
\label{app:decomposition}

For Stage~3 (Compositional StateTree), each target question is decomposed into a $2 \times 2 \times 2$ hierarchy using an LLM. Given a target question such as ``When did Jon start expanding his studio's social media presence?'', the decomposition produces three step fragments along semantic axes:

\begin{itemize}[nosep,leftmargin=11pt]
    \item \textbf{Step 1} (Person selection): \texttt{[A] Jon}
    \item \textbf{Step 2} (Event category): \texttt{[B] expanding his studio's social media presence}
    \item \textbf{Step 3} (Question detail): \texttt{When did [A] start [B]?}
\end{itemize}

The full decomposition tree for one conversation contains 8 such paths (one per leaf), each branching on person at level 1, event category at level 2, and question specificity at level 3. Table~\ref{tab:decompose_example} shows the structure of a complete decomposition tree.

\begin{table}[h]
\centering
\small
\caption{Semantic decomposition example for the target question ``\textit{When did Jon start expanding his studio's social media presence?}'' (answer: ``April, 2023''). $\bigstar$ marks the target leaf. The correct path aggregates: step\_1\,=\,``[A] Jon'', step\_2\,=\,``[B] expanding his studio's social media presence'', step\_3\,=\,``When did [A] start [B]?'', yielding the full question after substitution.}
\label{tab:decompose_example}
\begin{tabular}{@{}p{0.08\textwidth}p{0.42\textwidth}p{0.43\textwidth}@{}}
\toprule
\textbf{Level 1} & \textbf{Level 2 (Event)} & \textbf{Level 3 (Question)} \\
\midrule
\multirow{4}{*}{\makecell[l]{[A] Jon}}
  & \multirow{2}{*}{[B] expanding his studio's social media presence}
    & $\bigstar$ \textit{When did [A] start [B]?} \\
  & & \textit{What did [A] do while [B]?} \\
\cmidrule(l){2-3}
  & \multirow{2}{*}{[B] going to a fair for exposure}
    & \textit{When did [A] start [B]?} \\
  & & \textit{What events did [A] participate in while [B]?} \\
\midrule
\multirow{4}{*}{\makecell[l]{[A] Gina}}
  & \multirow{2}{*}{[B] launching an ad campaign}
    & \textit{When did [A] start [B]?} \\
  & & \textit{Why did [A] decide to start [B]?} \\
\cmidrule(l){2-3}
  & \multirow{2}{*}{[B] teaming up with a local artist}
    & \textit{When did [A] start [B]?} \\
  & & \textit{What did [A] do while [B]?} \\
\bottomrule
\end{tabular}
\end{table}

\subsubsection{Decomposition LLM Prompt}
\label{app:decomp_prompt}

To generate the $2 \times 2 \times 2$ decomposition trees for Stage~3, we use GPT-4o with temperature 0.7. Each target question is processed with the following system and user prompts. The system prompt specifies the tree structure, output schema, and quality constraints; the user prompt provides the target question along with the full question pool for that conversation.

\begin{PromptBox}{Decomposition System Prompt (Stage~3 Data Construction)}
You are a dataset construction assistant. Your task is to decompose a target question into a $2\times2\times2$ binary discrimination tree for a reading comprehension benchmark.

\par
The tree has 3 levels:
\begin{itemize}[nosep,leftmargin=1em]
\item Level 1 (root $\to$ 2 branches): Person selection. Each branch carries step\_1 = ``[A] PersonName''
\item Level 2 (each person $\to$ 2 branches): Event category. Each branch carries step\_2 = ``[B] gerund\_phrase''. The gerund phrase must be concrete and specific (e.g., ``attending the LGBTQ support group''), NOT a vague category (e.g., ``community activities'').
\item Level 3 (each event $\to$ 2 branches): Question specificity. Each branch carries step\_3 = a question template that MUST contain BOTH [A] and [B] as placeholders. The two questions under the same event MUST ask about different factual aspects.
\end{itemize}

\par
CRITICAL: step\_3 must ask about CONCRETE FACTS that have definite answers --- things like time, place, people involved, specific actions taken, specific results/outcomes, or specific objects.

GOOD examples (factual): ``When did [A] [B]?'', ``What did [A] make while [B]?'', ``Where did [A] go for [B]?''

BAD examples (subjective/generic): ``How did [A] feel about [B]?'', ``Why did [A] enjoy [B]?''

\par
Exactly ONE of the 8 leaves must be marked is\_target=true --- the leaf whose step\_3, after substituting [A] and [B], yields (or is semantically equivalent to) the original target question.

\par
Rules:
\begin{enumerate}[nosep,leftmargin=1.5em]
\item The target person MUST appear first in the tree.
\item The target event MUST appear first under the target person.
\item The target leaf MUST be the first leaf under the target event.
\item Each event\_gerund should be a present participle phrase derived from an actual question in the pool.
\item Every leaf's ``answer'' field MUST be copied verbatim from a question in the pool.
\item The two leaves under the same event MUST ask about genuinely DIFFERENT ASPECTS of the event.
\item Events for the same person should be distinct and both grounded in the question pool.
\item Events for the other person should also correspond to real pool questions but thematically contrast with the target person's events.
\item All answers must be short (a few words or a short phrase), copied from the pool.
\item EVERY step\_3 MUST contain both [A] and [B] placeholders.
\end{enumerate}
\end{PromptBox}

\begin{PromptBox}{Decomposition User Prompt Template (Stage~3 Data Construction)}
Target question (row \{row\_index\}):
\par
\quad Q: \{target\_question\}
\par
\quad sample\_id: \{sample\_id\}

\par
Speakers in this conversation:
\par
\quad Speaker A: \{speaker\_a\}
\par
\quad Speaker B: \{speaker\_b\}

\par
Below is the full question pool for this conversation (\{num\_questions\} questions). Use these to find suitable distractor questions and events. Pick events and questions that are grounded in the actual conversation content.

\par
--- Question Pool ---
\par
\{question\_pool\}
\par
--- End of Question Pool ---

\par
Please generate the $2\times2\times2$ decomposition tree JSON for this target question. Remember:
\begin{itemize}[nosep,leftmargin=1em]
\item step\_2 must describe the EVENT/ACTIVITY itself, NOT reveal or hint at the answer
\item EVERY step\_3 MUST contain BOTH [A] and [B] placeholders
\item step\_3 must ask about CONCRETE FACTS
\item The target person and target event come first in the tree
\item Exactly 1 leaf is is\_target=true
\item Output raw JSON only, no markdown fencing
\end{itemize}
\end{PromptBox}

The question pool for each conversation is constructed by collecting all non-adversarial QA pairs associated with that conversation in the LoCoMo dataset, including their answers and evidence session numbers. Each generated decomposition is validated against a set of structural constraints (correct number of persons, events, leaves; presence of [A]/[B] placeholders; exactly one target leaf; semantic diversity of sibling step\_3 values) and regenerated if validation fails.

\subsection{Distractor Question Pool}
\label{app:distractor_pool}

At each StateTree stage (1--3), non-target leaves are populated with \emph{distractor questions} from the same conversation and training split (excluding the target). Their semantic relevance to the dialogue makes them strong distractors. For Stage~3, distractors are generated by the $2 \times 2 \times 2$ decomposition (Appendix~\ref{app:decomposition}), with each non-target leaf differing in at least one semantic axis.

\subsection{Record Insertion Strategy}
\label{app:insertion}

Tree edge records must be embedded naturally within the conversation text so that they blend into the dialogue flow. We employ the following insertion strategy:

\textbf{Insertion positions.}
For all tree-augmented stages (Stages~1--3), records are inserted \emph{inside speaker quotations}: within lines of the form \texttt{Speaker said, "..."}, records are placed at sentence boundaries (after periods, exclamation marks, or question marks) inside the quoted text. This prevents artifacts such as a JSON record appearing immediately before a \texttt{DATE:} line or between speaker turns.

\textbf{Balanced session distribution.}
Records are distributed across conversation sessions using a least-loaded-first allocation strategy. For each record to be inserted, we:
\begin{enumerate}[nosep,leftmargin=1.5em]
    \item Group available insertion slots by their enclosing session.
    \item Identify the session(s) with the fewest records already assigned.
    \item Randomly select a slot from the least-loaded session(s).
\end{enumerate}

\textbf{Same-key separation constraint.}
When the same UUID key appears in two records (as required by the temporal discrimination mechanism), these two records are guaranteed to be inserted into \emph{different} sessions. This is enforced by tracking which sessions have already been used for each key and excluding them from the candidate set.

\textbf{Post-insertion temporal ordering.}
After all records are inserted, a verification pass checks each temporal discrimination fork: the correct-path record must reside in a strictly more recent session (by parsed \texttt{DATE:} timestamp) than its distractor counterpart. If any violation is detected, the two records' positions in the context are swapped. This swap procedure ensures the temporal ordering invariant holds regardless of the initial random placement.

\subsection{Entity Key Pool for Ablation}
\label{app:entity_keys}

In the \emph{w/ Entity Keys} ablation (Section~\ref{sec:ablation}), each UUID is replaced with a high-frequency concrete noun sampled without replacement from the 14-word pool below. This ensures that performance differences arise from semantic key content rather than topical relevance.

\begin{table}[h]
    \centering
    \caption{Entity key pool used in the \emph{w/ Entity Keys} ablation.}
    \label{tab:entity_key_pool}
    \begin{tabular}{lllllll}
        \toprule
        % 1 & 2 & 3 & 4 & 5 & 6 & 7 \\
        % \midrule
        garden   & mirror   & temple   & anchor   & lantern  & ribbon   & candle   \\
        shield   & bridge   & castle   & meadow   & feather  & saddle   & trumpet  \\
        \bottomrule
    \end{tabular}
\end{table}

For example, under $D{=}3$ (7 keys required), a random draw might produce the edge records:
\begin{center}
\small
\texttt{\{``garden'': ``bridge''\}, \{``garden'': ``meadow''\}, \{``bridge'': ``castle''\}, \ldots}
\end{center}
instead of the standard UUID format \texttt{\{``f391e945-\ldots'': ``8ada7089-\ldots''\}}. The key pairing mechanism is preserved, but the model can now rely on the semantic familiarity of the word rather than performing pure string matching.

\section{Full Prompt Templates}
\label{app:prompts}

\subsection{System Prompt}
\label{app:system_prompt}

The following system prompt is used across all curriculum stages. It instructs the model to produce a step-by-step reasoning trace enclosed in \verb|<think>...</think>| tags and place the final answer inside \verb|\boxed{...}|.

\begin{PromptBox}{System Prompt}
A conversation between User and Assistant. The User asks a question, and the Assistant solves it. The Assistant first thinks about the reasoning process in the mind and then provides the User with the answer. The reasoning process is enclosed within <think> </think> and answer is enclosed within \textbackslash boxed\{\} tags, respectively, i.e., <think> reasoning process here </think> \textbackslash boxed\{answer here\}.
\end{PromptBox}

\subsection{Warm-up: Direct Dialogue QA}

The Warm-up prompt is identical to the official LoCoMo evaluation prompt \cite{maharana2024evaluating}.

\begin{PromptBox}{Warm-up Prompt}
Based on the above conversations, write a short answer for the following question in a few words. Do not write complete and lengthy sentences. Answer with exact words from the conversations whenever possible.

\par
Question: \texttt{\{question\}}
\end{PromptBox}

\subsection{Stage~1 \& 2: Basic StateTree ($D{=}2$ / $D{=}3$)}

Stages~1 and 2 share the same prompt template; only the tree depth differs.

\begin{PromptBox}{Basic StateTree Prompt (Stages~1 \& 2)}
In the conversation above, there are JSON records like \texttt{\{``KEY'': ``VALUE''\}} scattered throughout the dialogue text. They form UUID chains: starting from a given key, each value either points to the next key (a UUID) or contains the final question to answer.

\par
Your task: start from key \texttt{``\{root\_uuid\}''} and follow the chain to find the question, then answer it.

\par
How to follow the chain:
\begin{enumerate}[nosep,leftmargin=1.5em]
\item Search the entire conversation for all records whose key matches the current UUID.
\item Each record sits inside a specific session. Look at the nearest preceding \texttt{``DATE: ...''} line to determine that record's time.
\item If the same key appears in multiple sessions, choose which record to use: among the remaining records, pick the one from the most recent session DATE.
\item Read the chosen value:
  \begin{itemize}[nosep,leftmargin=1em]
  \item If it is a UUID, treat it as the next key and go back to step 1.
  \item If it is a natural-language question, that is the question you must answer.
  \end{itemize}
\item Once you find the question, answer it based on the conversation content.
\end{enumerate}
\end{PromptBox}

\subsection{Stage~3: Compositional StateTree ($D{=}3$)}

\begin{PromptBox}{Compositional StateTree Prompt (Stage~3)}
In the conversation above, there are JSON records like \texttt{\{``UUID'': \{``step'': ``...'', ``next'': ``UUID''\}\}} scattered throughout the dialogue text. They form a decision tree.

\par
Each record maps a UUID key to a JSON object with:
\begin{itemize}[nosep,leftmargin=1em]
\item \texttt{``step''}: a piece of information to accumulate along the path
\item \texttt{``next''}: the UUID of the next node to follow (absent at leaf nodes)
\end{itemize}

\par
Your task: start from key \texttt{``\{root\_uuid\}''} and follow the tree to a leaf, accumulating the step information at each edge. Then answer the question.

\par
How to follow the tree:
\begin{enumerate}[nosep,leftmargin=1.5em]
\item Search the entire conversation for all records whose key matches the current UUID.
\item Each record sits inside a specific session. Look at the nearest preceding \texttt{``DATE: ...''} line to determine that record's session time.
\item If the same key appears in multiple sessions, pick the record from the most recent session DATE.
\item Read the \texttt{``step''} field and remember it.
\item If a \texttt{``next''} field exists, use it as the new key and go back to step 1.
\item If no \texttt{``next''} field exists, you have reached a leaf.
\end{enumerate}

\par
Once you reach a leaf, aggregate all the step information you collected along the path to form the final question. Then answer that question step by step based on the conversation content.
\end{PromptBox}

\subsection{CoT Prompt Baseline}
\label{app:cot_prompt}

For the \emph{w/ CoT prompt} ablation variant (Section~\ref{sec:ablation}), we modify the System Prompt to explicitly encourage step-by-step reasoning, without any StateTree structure or RL training. This baseline tests whether generic reasoning prompts can replicate the gains from explicit structural navigation.

\begin{PromptBox}{CoT System Prompt}
A conversation between User and Assistant. The User asks a question, and the Assistant solves it. The Assistant first thinks about the reasoning process \textbf{step by step} in the mind, carefully analyzing the question and searching through the conversation for relevant information. Then the Assistant provides the User with the answer. The reasoning process is enclosed within <think> </think> and answer is enclosed within \textbackslash boxed\{\} tags, respectively, i.e., <think> reasoning process here </think> \textbackslash boxed\{answer here\}.
\end{PromptBox}

% The user instruction remains identical to the Warm-up prompt (Appendix~\ref{app:prompts}).

\subsection{LLM Evaluation Prompts}
\label{app:llm_eval_prompts}

% We use LLM-as-a-Judge to evaluate answer correctness on LoCoMo and LongMemEval. The complete prompts are provided below.

\subsubsection{LoCoMo Evaluation Prompt}
\label{app:locomo_eval_prompt}

\begin{PromptBox}{LoCoMo LLM Judge Prompt}
Your task is to label an answer to a question as "CORRECT" or "WRONG".

\par

You will be given the following data: (1) a question (posed by one user to another user), (2) a 'gold' (ground truth) answer, (3) a generated answer which you will score as CORRECT/WRONG.

\par

The point of the question is to ask about something one user should know about the other user based on their prior conversations. The gold answer will usually be a concise and short answer that includes the referenced topic, for example:

\par

Question: Do you remember what I got the last time I went to Hawaii?
\par
Gold answer: A shell necklace

\par

The generated answer might be much longer, but you should be generous with your grading --- as long as it touches on the same topic as the gold answer, it should be counted as CORRECT.

\par

For time related questions, the gold answer will be a specific date, month, year, etc. The generated answer might be much longer or use relative time references (like 'last Tuesday' or 'next month'), but you should be generous with your grading --- as long as it refers to the same date or time period as the gold answer, it should be counted as CORRECT. Even if the format differs (e.g., 'May 7th' vs '7 May'), consider it CORRECT if it's the same date.

\par

Now it's time for the real question:

\par

Question: \naen{\texttt{\{question\}}}
\par
Gold answer: \naen{\texttt{\{gold\_answer\}}}
\par
Generated answer: \naen{\texttt{\{generated\_answer\}}}

\par

Return the label CORRECT or WRONG in a json format with the key as "label". Do NOT include both CORRECT and WRONG in your response, or it will break the evaluation script.
\end{PromptBox}

\subsubsection{LongMemEval Evaluation Prompts}
\label{app:longmemeval_eval_prompt}

LongMemEval contains multiple task categories, each with a dedicated judge prompt.

\begin{PromptBox}{LongMemEval Standard Tasks Prompt}
I will give you a question, a correct answer, and a response from a model. Please answer yes if the response contains the correct answer. Otherwise, answer no. If the response is equivalent to the correct answer or contains all the intermediate steps to get the correct answer, you should also answer yes. If the response only contains a subset of the information required by the answer, answer no.

\par

Question: \naen{\texttt{\{question\}}}

\par

Correct Answer: \naen{\texttt{\{answer\}}}

\par

Model Response: \naen{\texttt{\{response\}}}

\par

Is the model response correct? Answer yes or no only.
\end{PromptBox}

\begin{PromptBox}{LongMemEval Temporal Reasoning Prompt}
I will give you a question, a correct answer, and a response from a model. Please answer yes if the response contains the correct answer. Otherwise, answer no. If the response is equivalent to the correct answer or contains all the intermediate steps to get the correct answer, you should also answer yes. If the response only contains a subset of the information required by the answer, answer no. In addition, do not penalize off-by-one errors for the number of days. If the question asks for the number of days/weeks/months, etc., and the model makes off-by-one errors (e.g., predicting 19 days when the answer is 18), the model's response is still correct.

\par

Question: \naen{\texttt{\{question\}}}

\par

Correct Answer: \naen{\texttt{\{answer\}}}

\par

Model Response: \naen{\texttt{\{response\}}}

\par

Is the model response correct? Answer yes or no only.
\end{PromptBox}

\begin{PromptBox}{LongMemEval Knowledge Update Prompt}
I will give you a question, a correct answer, and a response from a model. Please answer yes if the response contains the correct answer. Otherwise, answer no. If the response contains some previous information along with an updated answer, the response should be considered as correct as long as the updated answer is the required answer.

\par

Question: \naen{\texttt{\{question\}}}

\par

Correct Answer: \naen{\texttt{\{answer\}}}

\par

Model Response: \naen{\texttt{\{response\}}}

\par

Is the model response correct? Answer yes or no only.
\end{PromptBox}

\begin{PromptBox}{LongMemEval Single-session Preference Prompt}
I will give you a question, a rubric for desired personalized response, and a response from a model. Please answer yes if the response satisfies the desired response. Otherwise, answer no. The model does not need to reflect all the points in the rubric. The response is correct as long as it recalls and utilizes the user's personal information correctly.

\par

Question: \naen{\texttt{\{question\}}}

\par

Rubric: \naen{\texttt{\{answer\}}}

\par

Model Response: \naen{\texttt{\{response\}}}

\par

Is the model response correct? Answer yes or no only.
\end{PromptBox}

\section{Examples of Inserted Records in Context}
\label{app:record_examples}

This section illustrates how tree records are embedded within conversation text at each stage.

\subsection{Basic StateTree Record (Stages~1 \& 2)}

\subsubsection{Case Study}
In the Basic StateTree, each record is a flat key-value pair \texttt{\{UUID: VALUE\}} inserted at a natural dialogue boundary. The value is either another UUID (internal node) or a natural-language question (leaf node).

\begin{PromptBox}{Basic StateTree: Record Inserted in Dialogue}
Caroline said, ``Thanks, Melanie!
\par
\textcolor{green!60!black}{\texttt{\{``95c49518-3f28-45f4-85be-de9a66ea5fd9'': ``What are Caroline's plans for the summer?''\}}}.
\par
I really appreciate it. Excited for the future! Bye!''
\end{PromptBox}

In this example, the UUID key \texttt{95c49518...} maps directly to a question string, indicating a leaf node. If this were an internal node, the value would be another UUID pointing to the next hop.

The example below illustrates a concrete instance with $D{=}2$.

\textcolor{green!60!black}{Green} marks the correct path to leaf (target question); \textcolor{red!70!black}{red} marks distractor path to distractor leaves.
\vspace{-1ex}

\begin{PromptBox}{Example of StateTree-augmented dialogue input (Basic $D{=}2$)}
% \begin{center}
%     \fontsize{8}{8} \selectfont
% \begin{tcolorbox}[%
%     colback=white,
%     colframe=gray!75!black,
%     title=Example of StateTree-augmented dialogue input (Basic $D{=}2$),
%     after upper=\vspace{-1mm},   % 缩小上半部分文字与分割线的间距
%     before lower=\vspace{-1mm}   % 缩小分割线与下半部分文字的间距
%     ]

    % \small\ttfamily
    Session 1 --- \textbf{2:24 pm, 14 Aug, 2023} \\
    \textbf{Caroline:} \textcolor{gray}{That pic is cool! Representing inclusivity and diversity in my art is important to me.} \textcolor{green!60!black}{\texttt{\{"113464e2-8d54-493c-874a-ebb3c07b9d48": "5ab90ad1-5990-4820-9aec-6525e8e2867e"\}}} \textcolor{gray}{I also use it to speak up for the community and push for acceptance. Here's a recent painting! \ldots} \\[2pt]
    Session 2 --- \textbf{1:56 pm, 8 May, 2023} \\
    \textbf{Melanie:} \textcolor{gray}{So cool you found such a helpful group. What's it done for you?} \textcolor{red!70!black}{\texttt{\{"113464e2-8d54-493c-874a-ebb3c07b9d48": "9320b92a-cb63-4ed7-b22e-d3521da21950"\}}} \textcolor{gray}{\ldots} \\[2pt]
    Session 3 --- \textbf{8:18 pm, 6 July, 2023} \\
    \textbf{Caroline:} \textcolor{gray}{Glad you have support, Caroline!}  \textcolor{green!60!black}{\texttt{\{"5ab90ad1-5990-4820-9aec-6525e8e2867e": "When did Melanie paint a sunrise?"\}}} \textcolor{gray}{How's it going? \ldots} \\[2pt]
    Session 4 --- \textbf{7:55 pm, 9 June, 2023} \\
    \textbf{Melanie:} \textcolor{gray}{Glad you have support, Caroline!}  \textcolor{red!70!black}{\texttt{\{"5ab90ad1-5990-4820-9aec-6525e8e2867e": "What pets does Melanie have?"\}}} \textcolor{gray}{Unconditional love \ldots} \\[2pt]
    Session 5 --- \textbf{1:50 pm, 17 Aug, 2023} \\
    \textbf{Caroline:} \textcolor{gray}{Sounds great, Mel!} \textcolor{red!70!black}{\texttt{\{"9320b92a-cb63-4ed7-b22e-d3521da21950": "What kind of pot did Mel and her kids make with clay?"\}}} \textcolor{gray}{We'll \ldots} \\[2pt]
    Session 6 --- \textbf{1:36 pm, 3 July, 2023} \\
    \textbf{Melanie:} \textcolor{gray}{Wow, Caroline! That's great! I just signed up for a pottery class yesterday. It's like therapy for me, letting me express myself and get creative.} \textcolor{red!70!black}{\texttt{\{"9320b92a-cb63-4ed7-b22e-d3521da21950": "Would Melanie go on another roadtrip soon?"\}}} \textcolor{gray}{Have you \ldots} \\[3pt]
    \textit{\textbf{Instruction:}} In the dialogue above, a correct question is hidden within key:value records encoded with UUID strings. Starting from key \textcolor{blue}{113464e2-8d54-493c-874a-ebb3c07b9d48}, locate each record, resolve forks by selecting the edge record in the most recent session, follow the path to a leaf, then answer the recovered question.
    \tcblower
    \textit{\textbf{Walkthrough.}} Starting from root key \textbf{113464e2-8d54-493c-874a-ebb3c07b9d48}, the model finds two records in Session~1 (Aug~14) and Session~2 (May~8). Session~1 is more recent, so it follows \textbf{5ab90ad1-5990-4820-9aec-6525e8e2867e}. Next, it finds records for \textbf{5ab90ad1-5990-4820-9aec-6525e8e2867e} in Session~3 (July~6) and Session~4 (June~9). Session~3 is more recent, leading to the leaf ``When did Melanie paint a sunrise?''. The model answers this question from the dialogue content.
% \end{tcolorbox}
% \vspace{-1ex}
% \end{center}
\end{PromptBox}

\subsection{Compositional StateTree Record (Stage~3)}

In the Compositional StateTree, each record is a nested JSON object carrying a \texttt{step} field (the semantic fragment) and optionally a \texttt{next} field (the UUID of the child node).

\begin{PromptBox}{Compositional StateTree: Records Inserted in Dialogue}
(Root node - Level 1, person selection)
\par
\ldots cherishing those happy moments and clinging to them is key.''
\par
\textcolor{green!60!black}{\texttt{\{``84858d3a-1f30-41cd-9882-afe161ca4970'': \{``step'': ``[A] Caroline'', ``next'': ``8ada7089-43a6-404c-b249-3d778e012633''\}\}}}.
\par
Melanie said, ``Yeah, same here Caroline.''

\par
(Internal node - Level 2, event selection)
\par
\ldots ``I've been thinking about volunteering more at the community center.''
\par
\textcolor{green!60!black}{\texttt{\{``8ada7089-43a6-404c-b249-3d778e012633'': \{``step'': ``[B] expanding his studio's social media presence'', ``next'': ``6916ce1a-c2d0-4146-9e8d-185110a68332''\}\}}}.
\par
Caroline said, ``That sounds like a great idea!''

\par
(Leaf node - Level 3, no ``next'' field)
\par
\ldots ``Reminds me it's important to cultivate a loving and accepting environment.''
\par
\textcolor{green!60!black}{\texttt{\{``6916ce1a-c2d0-4146-9e8d-185110a68332'': \{``step'': ``When did [A] start [B]?''\}\}}}.
\par
and shared a photo of a group of people\ldots
\end{PromptBox}

The root record provides \texttt{step: "[A] Caroline"} (person selection), the intermediate record provides \texttt{step: "[B] expanding his studio's social media presence"} (event selection), and the leaf record provides \texttt{step: "When did [A] start [B]?"} (question detail) with no \texttt{next} field. By following the correct path and aggregating all steps, the model reconstructs the full question.

\section{Hyperparameters and Training Configuration}
\label{app:hyperparams}

\subsection{Group Relative Policy Optimization (GRPO)}
\label{sec:grpo}

We adopt GRPO to train our model. For each question $q$, its long context $\mathcal{L}$ (the StateTree-augmented dialogue $\mathcal{S}'$ defined in Section~\ref{sec:data_construction_pipeline}), and its ground-truth answer $a$ from a dataset $\mathcal{D}$, GRPO samples a group of rollout trajectories $\{o_1,o_2,\cdots,o_{G}\}$ from the old policy $\pi_{\theta_{old}}$. The policy $\pi_\theta$ is then optimized by maximizing:
\begin{align}
	\!\!\!J_{\text{GRPO}}(\theta) 
	&= \mathbb{E}_{(\mathcal{L}, q, a) \sim \mathcal{D},\, \{o_i\}_{i=1}^{G} \sim \pi_{\theta_{\text{old}}}(\cdot|\mathcal{L}, q)} \notag \\
	\Bigg[ \frac{1}{G} \!  \sum_{i=1}^{G} \! \frac{1}{|o_i|} \! \sum_{t=1}^{|o_i|} \!
	& \Big( \! \min \big[ \rho_{i,t}(\theta) A_i, 
	\mathrm{clip}(\rho_{i,t}(\theta), 1\!-\!\varepsilon, 1\!+\!\varepsilon) A_i \big]
	\!- \! \beta D_{\text{KL}}\big(\pi_\theta(\cdot|q,o_{i,<t}) \| \pi_{\text{ref}}(\cdot|q,o_{i,<t})\big) \Big) \Bigg]\!\!
\end{align}
where $\rho_{i,t}(\theta) = \frac{\pi_\theta(o_{i,t} | q, o_{i,<t})}{\pi_{\theta_{\text{old}}}(o_{i,t} | q, o_{i,<t})}$. Hyperparameters $\varepsilon$ and $\beta$ control the clipping range of the importance sampling ratio and the weight of the KL penalty term, respectively. The advantage $A_i$ is computed by normalizing the trajectory rewards $\{r_1,r_2, \dots, r_G\}$ for each rollout trajectory:
\begin{align}
A_i=\frac{r_i-\text{mean}(\{r_1,r_2,\cdots,r_G\})}{\text{std}(\{r_1,r_2,\cdots,r_G\})}
\end{align}
Here, $r_i$ is the reward for trajectory $o_i$. Following \citep{guo2025deepseek, kimi1.5}, we use a combined reward function to assign $r_i$ and mitigate reward hacking.

Training hyperparameters are provided in Section~\ref{sec:rl} of the main text. Each experiment is run three times with different random seeds, and we report the average results across runs.

\subsection{Compute Resources}

All training runs are performed on an internal cluster with 32$\times$NVIDIA H20 GPUs (96~GB). The training times are approximately 90~hours for Qwen2.5-7B-Instruct, 100~hours for Qwen3-8B, and 160~hours for Qwen2.5-14B-Instruct.

\section{Evaluation Details}
\label{app:eval}

\subsection{Benchmark Descriptions}

\begin{itemize}[leftmargin=11pt]
    \item \textbf{LoCoMo}~\citep{maharana2024evaluating}: A multi-session dialogue benchmark with four reasoning categories: Multi Hop, Temporal, Open Domain, and Single Hop. Input lengths are approximately 10K tokens. We report Accuracy, token-level F1, and BLEU-1.

    % \item \textbf{LongMemEval}~\citep{wu2025longmemeval}: An out-of-domain benchmark evaluating six dimensions of long-term memory: Temporal reasoning, Multi-Session reasoning, Knowledge Update, and three Single-Session categories (User, Assistant, Preference). Input lengths range from 16K to 100K tokens.
    \item \textbf{LongMemEval}~\citep{wu2025longmemeval}: An out-of-domain benchmark evaluating six dimensions of long-term memory: Temporal reasoning, Multi-Session reasoning, Knowledge Update, and three Single-Session categories (User, Assistant, Preference). Input lengths range from 16K to 100K tokens. For readability, Table~\ref{tab:ood_results} abbreviates these categories as Temp., Multi-Ses., Know. Upd., SS-User, SS-Asst., and SS-Pref., respectively.

    \item \textbf{MMLU}~\citep{hendrycks2020mmlu}: A multiple-choice benchmark covering 57 academic subjects, testing general knowledge and reasoning.

    \item \textbf{MATH-500}~\citep{math-500}: A curated subset of 500 competition-level mathematics problems.

    \item \textbf{IFEval}~\citep{zhou2023instruction}: An instruction-following evaluation measuring the model's ability to comply with specific formatting and content constraints.

    % \item \textbf{PersonaMem}~\citep{jiang2025know}: A benchmark for evaluating dynamic user profiling and personalized responses at scale, measuring the model's ability to track preference evolution, recall shared facts, and provide preference-aligned recommendations across multi-turn conversations. Input lengths range from 32K to 128K tokens. We report exact-match accuracy.
    \item \textbf{PersonaMem}~\citep{jiang2025know}: A benchmark for evaluating dynamic user profiling and personalized responses at scale, measuring the model's ability to track preference evolution, recall shared facts, and provide preference-aligned recommendations across multi-turn conversations. Input lengths range from 32K to 128K tokens. We report exact-match accuracy. Table~\ref{tab:ood_results} abbreviates the seven task dimensions as Latest Pref.~(Acknowledge Latest User Preferences), New Scen.~(Generalize to New Scenarios), Align. Rec.~(Provide Preference-Aligned Recommendations), Shared Fact~(Recall User Shared Facts), Revisit Reas.~(Revisit Reasons Behind Preference Updates), New Idea~(Suggest New Ideas), and Track Evol.~(Track Full Preference Evolution).

\end{itemize}

\subsection{Inference Configuration}

For all evaluations, we use temperature 0.6, maximum input length of 128K tokens, and maximum output length of 4096 tokens.

\subsection{Additional OOD Results}

On PersonaMem 32k (Table~\ref{tab:model_comparison_3lines}), \lib-14B achieves 64.52\% average accuracy (+5.61\% over base), with particularly strong performance on recalling update reasons (86.87\%), aligning with the StateTree's explicit tracking of state transitions.

\input{tab/personamem}

\section{Category-Level Curriculum Ablation}
\label{app:category_ablation}

Tables~\ref{tab:locomo_category_ablation} and~\ref{tab:longmemeval_category_ablation} report the per-category breakdown of the curriculum stage ablation on LoCoMo and LongMemEval, respectively.

\begin{table}[h]
    \centering
    \caption{Category-level curriculum ablation on LoCoMo (7B).}
    \label{tab:locomo_category_ablation}
    \resizebox{0.8\textwidth}{!}{
    \begin{tabular}{lccccc}
        \toprule
        \textbf{Variant} & \textbf{Multi Hop} & \textbf{Temporal} & \textbf{Open Domain} & \textbf{Single Hop} & \textbf{Average} \\
        \midrule
        Qwen2.5-7B-Instruct & 63.77 & 23.90 & 50.00 & 44.92 & 44.29 \\
        \midrule
        \lib-7B (full) & 64.49 & 32.70 & 56.00 & 50.35 & 49.61 \\
        w/o warm-up & 62.32 & 27.67 & 50.00 & 45.63 & 45.19 \\
        w/o Stage~1 ($D{=}2$) & 63.04 & 24.53 & 50.00 & 45.63 & 44.68 \\
        w/o Stage~2 ($D{=}3$) & 60.14 & 28.93 & 52.00 & 47.52 & 46.23 \\
        w/o Stage~3 (Comp.) & 60.87 & 29.56 & 52.00 & 47.99 & 46.75 \\
        \bottomrule
    \end{tabular}
    }
\end{table}

\begin{table*}[t]
    \centering
    \caption{Category-level curriculum ablation on LongMemEval (7B).}
    \label{tab:longmemeval_category_ablation}
    \resizebox{\textwidth}{!}{
    \setlength{\tabcolsep}{3pt}
    \begin{tabular}{lccccccc}
        \toprule
        \textbf{Variant} & \textbf{Temporal} & \textbf{Multi-Session} & \textbf{Knowledge Update} & \textbf{Single-Session User} & \textbf{Single-Session Assistant} & \textbf{Single-Session Preference} & \textbf{Average} \\
        \midrule
        Qwen2.5-7B-Instruct & 20.30 & 8.27 & 52.56 & 25.71 & 37.50 & 3.33 & 23.80 \\
        \midrule
        \lib-7B (full) & 29.32 & 31.58 & 67.95 & 75.71 & 76.79 & 23.33 & 47.40 \\
        w/o warm-up & 27.82 & 28.57 & 64.10 & 68.57 & 67.86 & 13.33 & 43.00 \\
        w/o Stage~1 ($D{=}2$) & 24.81 & 25.56 & 62.82 & 71.43 & 73.21 & 20.00 & 42.60 \\
        w/o Stage~2 ($D{=}3$) & 26.32 & 28.57 & 64.10 & 72.86 & 73.21 & 20.00 & 44.20 \\
        w/o Stage~3 (Comp.) & 27.07 & 28.57 & 65.38 & 72.86 & 75.00 & 20.00 & 44.80 \\
        \bottomrule
    \end{tabular}
    }
\end{table*}

\textbf{Stage-wise contributions.}
Warm-up delivers the largest gains on Open Domain, Single-Session User/Assistant/Preference, and foundational QA.
Stage~1 ($D{=}2$) dominates Temporal and Multi-Session reasoning, indicating that shallow cross-session discrimination is the minimal effective structure for these capabilities.
Stage~2 ($D{=}3$) provides the largest Multi-Hop gains on LoCoMo, while Stage~3 (Compositional) contributes more uniformly across categories, consolidating prior skills through semantic decomposition.
In aggregate, the contribution rank order is Stage~1 $>$ Warm-up $>$ Stage~2 $>$ Stage~3 on both benchmarks (LoCoMo: +4.93 / +4.42 / +3.38 / +2.86; LongMemEval: +4.80 / +4.40 / +3.20 / +2.60), confirming that shallow temporal discrimination and QA grounding are the most critical components, with deeper compositional stages yielding diminishing but appreciable marginal returns.

\section{Baseline Implementation Details}
\label{app:baseline_details}

To ensure a fair comparison, we clarify the evaluation setup for each baseline:
\begin{itemize}[leftmargin=11pt]
    \item \textbf{SEALONG}~\citep{li2024large}: We evaluate using the officially released checkpoints provided by the authors.
    \item \textbf{RL-MemAgent}~\citep{yu2026memagent}: We evaluate using the officially released checkpoints provided by the authors.
    \item \textbf{LoongRL}~\citep{wang2026loongrl}: Since their model checkpoints are not publicly available, we re-train their method using their publicly released dataset on the Qwen2.5 base models, following their published training configuration.
\end{itemize}

\section{Complete Training Examples}
\label{app:training_examples}

This section shows training inputs for LoCoMo sample \texttt{conv-26} (``What is Caroline's identity?''). Only turns containing tree records are shown.

\subsection{Basic StateTree Training}
\label{app:l2_training}

The following Stage~2 sample uses flat \texttt{\{UUID: VALUE\}} records embedded in speaker quotations, where values are either UUIDs (internal nodes) or questions (leaves).

\begin{PromptBox}{Basic StateTree Training Input}
DATE: 2:24 pm on 14 August, 2023

\par

Caroline said, ``It was so inspiring, Mel! Check out the crowd. \textcolor{green!60!black}{\texttt{\{\textquotedbl{}387c1508-38e1-43f0-a36d-ce8cba77a4c9\textquotedbl{}: \textquotedbl{}37cbcf3a-944f-45cd-b647-e5e12f51d593\textquotedbl{}\}}}. People of all kinds celebrating love and acceptance - it really pushed me to keep fighting for LGBTQ rights.''

\par

DATE: 4:33 pm on 12 July, 2023

\par

Melanie said, ``Caroline, so glad you got the support! Your experience really brought you to where you need to be. You're gonna make a huge difference! \textcolor{red!70!black}{\texttt{\{\textquotedbl{}77376180-3363-4fa3-95d7-f21859335c9e\textquotedbl{}: \textquotedbl{}What does Melanie do to destress?\textquotedbl{}\}}}. This book I read last year reminds me to always pursue my dreams, just like you are doing!''

\par

DATE: 8:18 pm on 6 July, 2023

\par

Melanie said, ``Glad you have support, Caroline! Unconditional love is so important. Here's a pic of my family camping at the beach. \textcolor{green!60!black}{\texttt{\{\textquotedbl{}37cbcf3a-944f-45cd-b647-e5e12f51d593\textquotedbl{}: \textquotedbl{}What is Caroline's identity?\textquotedbl{}\}}}. We love it, it brings us closer!''

\par

DATE: 1:36 pm on 3 July, 2023

\par

Melanie said, ``Wow, Caroline, sounds like the parade was an awesome experience! \textcolor{red!70!black}{\texttt{\{\textquotedbl{}37cbcf3a-944f-45cd-b647-e5e12f51d593\textquotedbl{}: \textquotedbl{}What was the poetry reading that Caroline attended about?\textquotedbl{}\}}}. It's great to see the love and support for the LGBTQ+ community. Congrats! Has this experience influenced your goals at all?''

\par

DATE: 1:14 pm on 25 May, 2023

\par

Caroline said, ``I chose them 'cause they help LGBTQ+ folks with adoption. Their inclusivity and support really spoke to me.\textcolor{red!70!black}{\texttt{\{\textquotedbl{}387c1508-38e1-43f0-a36d-ce8cba77a4c9\textquotedbl{}: \textquotedbl{}77376180-3363-4fa3-95d7-f21859335c9e\textquotedbl{}\}}}.''

\par

DATE: 1:56 pm on 8 May, 2023

\par

Melanie said, ``That's really cool. \textcolor{red!70!black}{\texttt{\{\textquotedbl{}77376180-3363-4fa3-95d7-f21859335c9e\textquotedbl{}: \textquotedbl{}When did Melanie's friend adopt a child?\textquotedbl{}\}}}. You've got guts. What now?''

In the conversation above, there are JSON records like {"KEY": "VALUE"} scattered throughout the dialogue text. They form UUID chains: starting from a given key, each value either points to the next key (a UUID) or contains the final question to answer.

Your task: start from key "387c1508-38e1-43f0-a36d-ce8cba77a4c9" and follow the chain to find the question, then answer it.

How to follow the chain:

1. Search the entire conversation for all records whose key matches the current UUID.

2. Each record sits inside a specific session. Look at the nearest preceding "DATE: ..." line to determine that record's time.

3. If the same key appears in multiple sessions, choose which record to use:

   - Among the remaining records, pick the one from the most recent session DATE.
   
4. Read the chosen value:

   - If it is a UUID, treat it as the next key and go back to step 1.
   
   - If it is a natural-language question, that is the question you must answer.
   
5. Once you find the question, answer it based on the conversation content.
\end{PromptBox}

\textbf{Chain head UUID:} 387c1508-38e1-43f0-a36d-ce8cba77a4c9 \\
\textbf{Question:} What is Caroline's identity? \\
\textbf{Answer:} Transgender woman

\textbf{Correct path} (2 edges):
\begin{enumerate}[nosep,leftmargin=1.5em]
    \item \texttt{\{``387c1508-38e1-43f0-a36d-ce8cba77a4c9'': ``37cbcf3a-944f-45cd-b647-e5e12f51d593''\}} [SESSION: 2:24 pm on 14 August, 2023]
    \item \texttt{\{``37cbcf3a-944f-45cd-b647-e5e12f51d593'': ``What is Caroline's identity?''\}} [SESSION: 8:18 pm on 6 July, 2023]
\end{enumerate}

\textbf{Temporal discrimination forks} (correct-path edges are the most recent among records sharing the same key):
\begin{itemize}[nosep,leftmargin=1em]
    \item \textbf{Depth 0} (UUID \texttt{387c1508-38e1-43f0-a36d-ce8cba77a4c9}):
    \begin{itemize}[nosep,leftmargin=1em]
        \item Correct edge: \texttt{37cbcf3a-944f-45cd-b647-e5e12f51d593} [2:24 pm on 14 August, 2023]
        \item Distractor edge: \texttt{77376180-3363-4fa3-95d7-f21859335c9e} [1:14 pm on 25 May, 2023]
    \end{itemize}
    \item \textbf{Depth 1} (UUID \texttt{37cbcf3a-944f-45cd-b647-e5e12f51d593}):
    \begin{itemize}[nosep,leftmargin=1em]
        \item Correct edge: ``What is Caroline's identity?'' [8:18 pm on 6 July, 2023]
        \item Distractor edge: ``What was the poetry reading that Caroline attended about?'' [1:36 pm on 3 July, 2023]
    \end{itemize}
\end{itemize}

\begin{table}[h]
\centering
\small
\caption{StateTree structure for the $D{=}2$ Basic StateTree sample (conv-26). $\bigstar$ marks the target leaf.}
\label{tab:l2_tree_structure}
\begin{tabular}{@{}p{0.20\textwidth}p{0.22\textwidth}p{0.5\textwidth}@{}}
\toprule
\textbf{Level 1 (Root)} & \textbf{Level 2 (Next)} & \textbf{Level 3 (Question)} \\
\midrule
% \multirow{4}{*}{\makecell[l]{\texttt{387c1508-38e1-43f0-a36d-ce8cba77a4c9}}}
\multirow{4}{*}{\makecell[l]{\texttt{387c1508-38e1-43f0-} \\ \texttt{a36d-ce8cba77a4c9}}}
  & \multirow{2}{*}{\makecell[l]{\texttt{37cbcf3a-944f-45cd-} \\ \texttt{b647-e5e12f51d593}}}
    & $\bigstar$ What is Caroline's identity? \\
  & & What was the poetry reading that Caroline attended about? \\
\cmidrule(l){2-3}
  & \multirow{2}{*}{\makecell[l]{\texttt{77376180-3363-4fa3-} \\ \texttt{95d7-f21859335c9e}}}
    & What does Melanie do to destress? \\
  & & When did Melanie's friend adopt a child? \\
\bottomrule
\end{tabular}
\end{table}

The 6 edge records are distributed across 6 distinct sessions (one per session).

\subsection{Compositional StateTree Training (Stage~3)}
\label{app:h3_training}

The following Stage~3 sample uses nested \texttt{\{UUID: \{``step'', ``next''\}\}} records in speaker quotations, where \texttt{step} carries a semantic fragment and \texttt{next} points to the child UUID (absent at leaves).

\begin{PromptBox}{Compositional StateTree Training Input}
DATE: 3:31 pm on 23 August, 2023

\par

Caroline said, ``Thanks, Melanie! I really appreciate it. Excited for the future! \textcolor{green!60!black}{\texttt{\{\textquotedbl{}4c6f1497-b881-46a7-8cc4-aba5fbac5429\textquotedbl{}: \{\textquotedbl{}step\textquotedbl{}: \textquotedbl{}[A] Caroline\textquotedbl{}, \textquotedbl{}next\textquotedbl{}: \textquotedbl{}1616ef6c-b7ed-4908-bbf4-cfa532591f0c\textquotedbl{}\}\}}}. Bye!''

\par

DATE: 1:50 pm on 17 August, 2023

\par

Caroline said, ``Sure thing, Melanie! \textcolor{red!70!black}{\texttt{\{\textquotedbl{}97ae53c4-dd29-4d61-bf87-140b35e082ac\textquotedbl{}: \{\textquotedbl{}step\textquotedbl{}: \textquotedbl{}[B] running a charity race\textquotedbl{}, \textquotedbl{}next\textquotedbl{}: \textquotedbl{}59b5b99c-03f2-4abc-8331-cfb5119c720b\textquotedbl{}\}\}}}. Can't wait to see your pottery project.  I'm happy you found something that makes you happy. Show me when you can!''

\par

DATE: 2:24 pm on 14 August, 2023

\par

Melanie said, ``Wow, Caroline, that's so cool! \textcolor{red!70!black}{\texttt{\{\textquotedbl{}4c6f1497-b881-46a7-8cc4-aba5fbac5429\textquotedbl{}: \{\textquotedbl{}step\textquotedbl{}: \textquotedbl{}[A] Melanie\textquotedbl{}, \textquotedbl{}next\textquotedbl{}: \textquotedbl{}97ae53c4-dd29-4d61-bf87-140b35e082ac\textquotedbl{}\}\}}}. Art can be so healing and a way to really connect with who you are. It's awesome that beauty can be found in the imperfections. We're all individual and wonderfully imperfect. Thanks for sharing it with me!''

\par

DATE: 8:56 pm on 20 July, 2023

\par

Caroline said, ``Sounds fun! What was the best part? Do you do it often with the kids?\textcolor{red!70!black}{\texttt{\{\textquotedbl{}59b5b99c-03f2-4abc-8331-cfb5119c720b\textquotedbl{}: \{\textquotedbl{}step\textquotedbl{}: \textquotedbl{}What did [B] that [A] was part of raise awareness for?\textquotedbl{}\}\}}}.''

\par

DATE: 2:31 pm on 17 July, 2023

\par

Melanie said, ``Wow, Caroline, that painting is awesome! Those colors are so vivid and the whole thing looks really unified. What inspired you?\textcolor{green!60!black}{\texttt{\{\textquotedbl{}1616ef6c-b7ed-4908-bbf4-cfa532591f0c\textquotedbl{}: \{\textquotedbl{}step\textquotedbl{}: \textquotedbl{}[B] discussing her identity\textquotedbl{}, \textquotedbl{}next\textquotedbl{}: \textquotedbl{}da0f6e17-39f3-460f-848a-eb3375327559\textquotedbl{}\}\}}}.''

\par

DATE: 1:51 pm on 15 July, 2023

\par

Caroline said, ``Wow, nice pic! \textcolor{red!70!black}{\texttt{\{\textquotedbl{}87801b39-e9da-4bac-a1a8-e272d4931144\textquotedbl{}: \{\textquotedbl{}step\textquotedbl{}: \textquotedbl{}What did [A] learn while [B]?\textquotedbl{}\}\}}}. You both looked amazing. One special memory for me was this pride parade I went to a few weeks ago.''

\par

DATE: 4:33 pm on 12 July, 2023

\par

Melanie said, ``Thanks, Caroline! \textcolor{red!70!black}{\texttt{\{\textquotedbl{}5fa0b1cb-87a6-4a2f-8a75-5db870313798\textquotedbl{}: \{\textquotedbl{}step\textquotedbl{}: \textquotedbl{}What types of paintings has [A] done related to [B]?\textquotedbl{}\}\}}}. This has been great for my mental health. I'm gonna keep it up.''

\par

DATE: 8:18 pm on 6 July, 2023

\par

Melanie said, ``Glad you have support, Caroline! Unconditional love is so important. Here's a pic of my family camping at the beach. \textcolor{red!70!black}{\texttt{\{\textquotedbl{}87801b39-e9da-4bac-a1a8-e272d4931144\textquotedbl{}: \{\textquotedbl{}step\textquotedbl{}: \textquotedbl{}When did [A] start [B]?\textquotedbl{}\}\}}}. We love it, it brings us closer!''

\par

DATE: 1:36 pm on 3 July, 2023

\par

Caroline said, ``Cool, thanks Mel! Can't wait. I'll keep ya posted. Bye!\textcolor{green!60!black}{\texttt{\{\textquotedbl{}da0f6e17-39f3-460f-848a-eb3375327559\textquotedbl{}: \{\textquotedbl{}step\textquotedbl{}: \textquotedbl{}What is [A]'s identity while [B]?\textquotedbl{}\}\}}}.''

\par

DATE: 10:37 am on 27 June, 2023

\par

Melanie said, ``Congrats Caroline! Good on you for going after what you really care about.\textcolor{red!70!black}{\texttt{\{\textquotedbl{}1616ef6c-b7ed-4908-bbf4-cfa532591f0c\textquotedbl{}: \{\textquotedbl{}step\textquotedbl{}: \textquotedbl{}[B] researching adoption agencies\textquotedbl{}, \textquotedbl{}next\textquotedbl{}: \textquotedbl{}87801b39-e9da-4bac-a1a8-e272d4931144\textquotedbl{}\}\}}}.''

\par

DATE: 7:55 pm on 9 June, 2023

\par

Melanie said, ``Yeah, Caroline! \textcolor{red!70!black}{\texttt{\{\textquotedbl{}97ae53c4-dd29-4d61-bf87-140b35e082ac\textquotedbl{}: \{\textquotedbl{}step\textquotedbl{}: \textquotedbl{}[B] painting a sunset\textquotedbl{}, \textquotedbl{}next\textquotedbl{}: \textquotedbl{}5fa0b1cb-87a6-4a2f-8a75-5db870313798\textquotedbl{}\}\}}}. It takes courage to talk about our own stories. But it's in these vulnerable moments that we bond and understand each other. We all have our different paths, but if we share them, we show people that they're not alone. Our stories can be so inspiring and encouraging to others who are facing the same challenges. Thank you for using your voice to create love, acceptance, and hope. You're doing amazing!''

\par

Melanie said, ``Absolutely, Caroline! \textcolor{red!70!black}{\texttt{\{\textquotedbl{}5fa0b1cb-87a6-4a2f-8a75-5db870313798\textquotedbl{}: \{\textquotedbl{}step\textquotedbl{}: \textquotedbl{}What did [A] paint while [B]?\textquotedbl{}\}\}}}. I cherish time with family. It's when I really feel alive and happy.''

\par

DATE: 1:14 pm on 25 May, 2023

\par

Melanie said, ``That's great, Caroline! Loving the inclusivity and support. \textcolor{red!70!black}{\texttt{\{\textquotedbl{}da0f6e17-39f3-460f-848a-eb3375327559\textquotedbl{}: \{\textquotedbl{}step\textquotedbl{}: \textquotedbl{}What career paths is [A] considering while [B]?\textquotedbl{}\}\}}}. Anything you're excited for in the adoption process?''

\par

DATE: 1:56 pm on 8 May, 2023

\par

Caroline said, ``I went to a LGBTQ support group yesterday and it was so powerful.\textcolor{red!70!black}{\texttt{\{\textquotedbl{}59b5b99c-03f2-4abc-8331-cfb5119c720b\textquotedbl{}: \{\textquotedbl{}step\textquotedbl{}: \textquotedbl{}When did [A] participate in [B]?\textquotedbl{}\}\}}}.''
\end{PromptBox}

\textbf{Root UUID:} 4c6f1497-b881-46a7-8cc4-aba5fbac5429 \\
\textbf{Original question:} What is Caroline's identity? \\
\textbf{Answer:} Transgender woman

\textbf{Correct path} (3 edges):
\begin{enumerate}[nosep,leftmargin=1.5em]
    \item Level 1: \texttt{\{``4c6f1497-b881-46a7-8cc4-aba5fbac5429'': \{``step'': ``[A] Caroline'', ``next'': ``1616ef6c-b7ed-4908-bbf4-cfa532591f0c''\}\}} [SESSION: 3:31 pm on 23 August, 2023]
    \item Level 2: \texttt{\{``1616ef6c-b7ed-4908-bbf4-cfa532591f0c'': \{``step'': ``[B] discussing her identity'', ``next'': ``da0f6e17-39f3-460f-848a-eb3375327559''\}\}} [SESSION: 2:31 pm on 17 July, 2023]
    \item Level 3: \texttt{\{``da0f6e17-39f3-460f-848a-eb3375327559'': \{``step'': ``What is [A]'s identity while [B]?''\}\}} [SESSION: 1:36 pm on 3 July, 2023]
\end{enumerate}

\begin{table}[h]
\centering
\small
\caption{Full $2 \times 2 \times 2$ tree structure for the Compositional StateTree sample (conv-26). $\bigstar$ marks the target leaf.}
\label{tab:h3_tree_structure}
\begin{tabular}{@{}p{0.12\textwidth}p{0.32\textwidth}p{0.5\textwidth}@{}}
\toprule
\textbf{Level 1} & \textbf{Level 2 (Event)} & \textbf{Level 3 (Question / Answer)} \\
\midrule
\multirow{4}{*}{\makecell[l]{[A] Caroline}}
  & \multirow{2}{*}{[B] discussing her identity (correct)} 
    & $\bigstar$ \textit{What is [A]'s identity while [B]?} $\rightarrow$ \texttt{Transgender woman} \\
  & & \textit{What career paths is [A] considering while [B]?} $\rightarrow$ \texttt{Psychology, counseling certification} \\
\cmidrule(l){2-3}
  & \multirow{2}{*}{[B] researching adoption agencies}
    & \textit{What did [A] learn while [B]?} $\rightarrow$ \texttt{Adoption agencies} \\
  & & \textit{When did [A] start [B]?} $\rightarrow$ \texttt{researching adoption agencies} \\
\midrule
\multirow{4}{*}{\makecell[l]{[A] Melanie}}
  & \multirow{2}{*}{[B] running a charity race}
    & \textit{When did [A] participate in [B]?} $\rightarrow$ \texttt{The sunday before 25 May 2023} \\
  & & \textit{What did [B] that [A] was part of raise awareness for?} $\rightarrow$ \texttt{mental health} \\
\cmidrule(l){2-3}
  & \multirow{2}{*}{[B] painting a sunset}
    & \textit{What did [A] paint while [B]?} $\rightarrow$ \texttt{sunset} \\
  & & \textit{What types of paintings has [A] done related to [B]?} $\rightarrow$ \texttt{Horse, sunset, sunrise} \\
\bottomrule
\end{tabular}
\end{table}

The 14 edge records are distributed across 8 distinct sessions.

\section{Qualitative Examples}
\label{app:qualitative}

% This section presents concrete examples of the four structured reasoning behaviors. 

\subsection{Cross-Session Retrieval}

The base model stops at the first matching session and returns a partial answer, whereas \lib enumerates all relevant records distributed across sessions before aggregating a complete answer.

\textbf{Question (LongMemEval):} ``How many fish are there in total in both of my aquariums?'' (Ground truth: \texttt{17})

\begin{PromptBox}{Qwen2.5-14B-Instruct Response}
From previous conversations, the user mentioned having a 10-gallon tank with a betta fish, Bubbles, and a 20-gallon tank. No specific number of fish in the 20-gallon tank was mentioned, only types like schooling fish. Therefore, the total number can only be estimated with the known fish in the 10-gallon tank.

\par
$\boxed{\text{1}}$
\end{PromptBox}

\begin{PromptBox}{\lib-14B Response}
From the conversation, the user mentioned having 10 neon tetras, 5 golden honey gouramis, and a small pleco catfish in their 20-gallon tank. They did not specify the number of fish in their other tank, but since the question is framed as if there is a known number in the other tank and based on previous context, we can infer they might be referring to their 10-gallon tank which previously had a betta fish named Bubbles. Assuming the 10-gallon tank has only Bubbles and no other fish added, and the 20-gallon tank has the mentioned 16 fish (10 + 5 + 1), the total number of fish would be 16 from the 20-gallon tank and 1 from the 10-gallon tank.

\par
So, the total number of fish in both tanks is 10 neon tetras + 5 golden honey gouramis + 1 pleco catfish + 1 betta fish = 16 + 1 = 17 fish.

\par
$\boxed{\text{17}}$
\end{PromptBox}

\textbf{Question (LongMemEval):} ``How many graduation ceremonies have I attended in the past three months?'' (Ground truth: \texttt{3})

\begin{PromptBox}{Qwen2.5-14B-Instruct Response}
The user mentioned attending a niece's preschool graduation ceremony on February 28th and a best friend's master's degree graduation ceremony a couple of weeks ago, which would be around March 10th to March 15th. So, the user has attended two graduation ceremonies in the past three months.

\par
$\boxed{\text{2}}$
\end{PromptBox}

\begin{PromptBox}{\lib-14B Response}
Based on the conversations, you have mentioned attending two graduation ceremonies in the past three months:
1. Your colleague Alex's graduation from a leadership development program at work a few weeks ago.
2. Your best friend Rachel's master's degree graduation ceremony last month from March 15th to 19th.
You also mentioned attending your little cousin Emma's preschool graduation about two months ago, and your niece's eighth grade graduation ceremony last month.

\par
So, the total number of graduation ceremonies you have attended in the past three months is 3.

\par
$\boxed{\text{3}}$
\end{PromptBox}

\subsection{Multi-Hop Reasoning}

The base model fails to compose individual facts into the required inference, whereas \lib chains scattered evidence into a coherent answer.

\textbf{Question (LongMemEval):} ``How old was I when Alex was born?'' (Ground truth: \texttt{11})

\begin{PromptBox}{Qwen2.5-14B-Instruct Response}
The user mentioned they are considering mentoring a 21-year-old intern named Alex. However, the user's age was not directly mentioned in relation to Alex's birth. Based on the available context, the user recently turned 32 and is considering a career change.

\par
$\boxed{\text{32}}$
\end{PromptBox}

\begin{PromptBox}{\lib-14B Response}
Based on the context provided, the user mentioned that they are 32 years old and are considering a career change for themselves. They also mentioned that they are a mentor to Alex, an intern who is 21 years old. However, the user did not directly state their age when Alex was born. Given that Alex is 21 years old and the user is 32, the user was 32 - 21 = 11 years old when Alex was born.

\par
$\boxed{\text{11}}$
\end{PromptBox}

\textbf{Question (LoCoMo):} ``When did John resume playing drums in his adulthood?'' (Ground truth: \texttt{February 2022})

\begin{PromptBox}{Qwen2.5-14B-Instruct Response}
From the conversation on 27 March, 2022, John mentioned he started playing drums a month before the conversation.

\par
$\boxed{\text{a month ago}}$
\end{PromptBox}

\begin{PromptBox}{\lib-14B Response}
From the conversation on 27 March, 2022, John mentioned that he had been playing drums for a month and that he started playing again recently. Since the conversation is dated 27 March, 2022, and John has been playing for a month by then, we can infer that he resumed playing drums sometime in February 2022.

\par
$\boxed{\text{February 2022}}$
\end{PromptBox}

\subsection{Temporal Reasoning}

The base model mis-orders events using superficial lexical cues, whereas \lib performs explicit chronological comparison via session timestamps.

\textbf{Question (LoCoMo):} ``Which activity did Sam resume in December 2023 after a long time?'' (Ground truth: \texttt{hiking})

\begin{PromptBox}{Qwen2.5-14B-Instruct Response}
Reviewing the conversations, there's no direct mention of Sam resuming any activity in December 2023. However, the closest relevant information is from October onwards where Sam mentions starting a new diet and exercise routine, and discussing plans for healthier living.

\par
$\boxed{\text{exercise routine}}$
\end{PromptBox}

\begin{PromptBox}{\lib-14B Response}
From the conversations, it is clear that Sam has been discussing and working on various activities such as dieting, exercising, and painting. However, the specific date of December 2023 is not mentioned directly. We need to infer from the context and timeline provided in the conversations.

\par
The conversations span from 6 June 2023 to 25 October 2023. Sam mentions starting a new diet and exercise routine and talks about his progress and setbacks. He also mentions that he had a health scare and a doctor's appointment that acted as a wake-up call to prioritize his health. Notably, Sam explicitly said ``I love hiking, but it's been ages since I've done it'' and shared a childhood memory of hiking with his dad. This ``it's been ages'' directly corresponds to the question's ``after a long time.'' Furthermore, in October 2023, Sam and Evan were actively planning a hike together---Sam expressed excitement, shared a photo of hiking shoes, and said ``Can't wait to go on a hike with you.'' Given that Sam had a long-standing interest in hiking but hadn't done it in a very long time, and was making concrete plans to resume it, it is logical to conclude that the activity Sam resumed after a long time is hiking.

\par
$\boxed{\text{hiking}}$
\end{PromptBox}

\textbf{Question (LongMemEval):} ``Which event did I attend first, the `Effective Time Management' workshop or the `Data Analysis using Python' webinar?'' (Ground truth: \texttt{`Data Analysis using Python' webinar})

\begin{PromptBox}{Qwen2.5-14B-Instruct Response}
The user mentioned attending a `Effective Time Management' workshop at the local community center last Saturday, and a `Data Analysis using Python' webinar two months ago. Based on the dates, the `Effective Time Management' workshop was attended first.

\par
$\boxed{\text{Effective Time Management workshop}}$
\end{PromptBox}

\begin{PromptBox}{\lib-14B Response}
Based on the conversations, the user mentioned attending the `Effective Time Management' workshop at the local community center on a specific date. The `Data Analysis using Python' webinar was mentioned as something they attended two months ago, which implies it was before the `Effective Time Management' workshop. Comparing the two: the webinar was attended two months prior to the workshop, placing it earlier on the timeline.

\par
$\boxed{\text{`Data Analysis using Python' webinar}}$
\end{PromptBox}

\subsection{Knowledge Update}

The base model conflates old and new mentions, retrieving outdated values, whereas \lib tracks the latest state by privileging more recent sessions.

\textbf{Question (PersonaMem):} ``I'm trying to decide whether to sign up for a structured painting class or just paint on my own time. Any advice?'' (Ground truth: \texttt{(b)})

\begin{itemize}[nosep,leftmargin=1em]
    \item[(a)] Considering you enjoy diving into new experiences and learning new things, a structured painting class might offer a fun and guided way to develop your skills. You can benefit from expert advice and meet new people who share your passion!
    \item[(b)] If you enjoy the freedom and creativity of spontaneous activities, you might find painting on your own time more fulfilling. This allows you to focus on quality and explore your artistic interests without the constraints of a structured class.
    \item[(c)] Painting on your own time gives you the chance to experiment with different styles and techniques at your own pace. It can be a more flexible option if you're looking for a stress-relieving activity that fits your schedule.
    \item[(d)] Both structured classes and painting on your own time have their merits. A class can provide valuable feedback and a social environment. If you appreciate structure, this might be the way to go.
\end{itemize}

\begin{PromptBox}{Qwen2.5-14B-Instruct Response}
Given the context of Alex's recent experiences, it appears that Alex values both structured and unstructured creative endeavors. However, Alex seems to prefer more casual and personal approaches, such as spontaneous movie nights and local film discussion groups, which suggest a preference for flexible and less stressful activities. Therefore, option (b) or (c) might resonate more with Alex's interests, but option (c) explicitly highlights the freedom and flexibility that aligns closely with Alex's desire for a stress-relieving activity that fits personal schedule.

\par
$\boxed{\text{(c)}}$
\end{PromptBox}

\begin{PromptBox}{\lib-14B Response}
Based on the context of Alex Rivera's preferences and experiences, it seems that Alex values both structured learning and personal freedom, but ultimately finds enjoyment and fulfillment in activities that are not overly stressful and allow for a more relaxed exploration of interests. Alex has mentioned stepping back from organized events that felt too structured or overwhelming, such as readathons and film festivals, to find more personal and fulfilling ways to engage with literature and film. However, Alex also expressed enthusiasm for organized events that foster community and learning, such as the local filmmaking workshop and the online film discussion group.

\par
Given this balance, Alex might appreciate the guidance and feedback from a structured painting class but also the flexibility and personal expression that comes from painting independently. However, the context leans slightly more towards the enjoyment of personal and less stressful activities, as Alex mentioned preferring spontaneous movie nights and smaller, more intimate film discussions over more structured events.

\par
Considering this, the most appropriate advice would be one that aligns with the value Alex places on personal fulfillment and less pressure. Option (b) emphasizes the freedom and creativity of painting on your own time, which seems to align well with Alex's preference for not feeling constrained by a rigid structure.

\par
$\boxed{\text{(b)}}$
\end{PromptBox}

\textbf{Question (LongMemEval):} ``How many new postcards have I added to my collection since I started collecting again?'' (Ground truth: \texttt{25})

\begin{PromptBox}{Qwen2.5-14B-Instruct Response}
Reviewing the conversation, the user mentioned adding 17 new postcards since they started collecting again.

\par
$\boxed{\text{17}}$
\end{PromptBox}

\begin{PromptBox}{\lib-14B Response}
From the conversation on 2023/05/20, the user mentioned they had added 17 new postcards since they started collecting again. On 2023/11/10, the user mentioned they had added 8 more postcards to their collection. However, the question specifically asks about the number of postcards added since they started collecting again, which initially was 17 postcards. Adding the 8 new postcards, the total number of new postcards added since the start of their collection is 25.

The exact phrase from the conversation was ``I've added 17 new ones since I started collecting again,'' and later ``I've added 25 new postcards to my collection since I started collecting again.''

Therefore, the total number of new postcards added since the start of the collection is 25.

\par
$\boxed{\text{25}}$
\end{PromptBox}

\textbf{Question (PersonaMem):} ``I came across a platform that offers various finance webinars, and I signed up.'' (Ground truth: \texttt{(b)})

\begin{itemize}[nosep,leftmargin=1em]
    \item[(a)] I remember you mentioned not enjoying finance-related webinars. It might be refreshing to explore something outside your comfort zone like these.
    \item[(b)] I remember you mentioned enjoying finance-related webinars. You might find them insightful for broadening your understanding of stock market strategies and financial planning.
    \item[(c)] I remember you mentioned enjoying travel-related webinars. You might find them insightful for broadening your understanding of global cultures and geography.
    \item[(d)] That's interesting! You must be excited to try out something new; how did you find that platform?
\end{itemize}

\begin{PromptBox}{Qwen2.5-14B-Instruct Response}
From the conversation, Alex mentioned not enjoying finance-related webinars due to their complexity and rapid information sharing, which can be overwhelming. However, since then, Alex has shown an increasing interest in financial literacy and has been exploring finance-related topics through various means such as reading, discussions, and personal projects. Given Alex's renewed engagement with finance, option (a) doesn't seem to fit as Alex is not looking to explore something outside their current interest. Option (c) is irrelevant since Alex has not expressed interest in travel-related webinars. Between (b) and (d), (d) seems more appropriate as it is more open-ended and does not assume Alex enjoys the webinars but rather acknowledges the newness of the platform and seeks feedback.

\par
$\boxed{\text{(d)}}$
\end{PromptBox}

\begin{PromptBox}{\lib-14B Response}
Based on the conversation, Alex mentioned that they had attended finance-related webinars and found them to be beneficial, even if some of the content was challenging. They expressed a positive attitude towards webinars in general, especially those related to finance. Therefore, the most appropriate response would be option (b), which aligns with Alex's past experiences and interests.

\par
$\boxed{\text{(b)}}$
\end{PromptBox}

%% file: tab/personamem.tex
\begin{table}[t]
    \centering
    \caption{Performance comparison across different models on OOD PersonaMem-32k.}
    \label{tab:model_comparison_3lines}
    
    % 使用 resizebox 确保表格宽度适应页面
    \resizebox{\textwidth}{!}{%
    \setlength{\tabcolsep}{3pt} % 列间距
    \begin{tabular}{lcccccccc}
        \toprule
        \textbf{Model} & 
        \makecell{\textbf{Generalizing} \\ \textbf{to New} \\ \textbf{Scenarios}} & 
        \makecell{\textbf{Provide Preference} \\ \textbf{Aligned} \\ \textbf{Recommendations}} & 
        \makecell{\textbf{Recall User} \\ \textbf{Shared} \\ \textbf{Facts}} & 
        \makecell{\textbf{Recalling Facts} \\ \textbf{Mentioned by} \\ \textbf{the User}} & 
        \makecell{\textbf{Recalling the Reasons} \\ \textbf{Behind Previous} \\ \textbf{Updates}} & 
        \makecell{\textbf{Suggest} \\ \textbf{New} \\ \textbf{Ideas}} & 
        \makecell{\textbf{Track Full} \\ \textbf{Preference} \\ \textbf{Evolution}} & 
        \textbf{Average} \\
        \midrule

        QwenLong-L1-32B & \textbf{71.93} & \textbf{65.45} & \textbf{67.44} & 52.94 & 78.79 & \textbf{22.58} & \textbf{74.82} & \textbf{63.84} \\
        R1-Distill-Qwen-32B  & 56.14 & 61.82 & 61.24 & \textbf{64.71} & \textbf{83.84} & \textbf{22.58} & 79.14 & 62.82 \\
        % R1-Distill-LLaMa-70B & 59.65 & 47.27 & \textbf{72.09} & 52.94 & \textbf{83.84} & 16.13 & \textbf{74.82} & 61.80 \\
        \midrule
        % --- 7B Models Group ---
        Qwen2.5-7B-Instruct      & 52.63 & 60.00 & 41.86 & 52.94 & 79.80 & 17.20 & 66.19 & 53.14 \\
        SEALONG-7B               & 56.14 & 54.55 & 41.09 & 47.06 & 82.83 & 17.20 & 64.75 & 52.80 \\
        LoongRL-7B               & 59.65 & \bf 69.09 & 48.84 & 64.71 & 78.79  & 20.43 & 72.66 & 58.40  \\
        RL-MemAgent-7B        & 52.63 & 49.09 & 47.29 & \bf 76.47 & 79.80 & \bf 34.41 & 57.55 & 54.67 \\
        \textbf{\lib-7B}   & \bf 68.42 & 63.64 & \bf 49.61 & \bf 76.47 & \bf 82.83 & 27.96 & \bf 73.38 & \bf 61.29  \\
        
        \midrule
        
        % --- 14B Models Group ---
        Qwen2.5-14B-Instruct     & \bf 75.44 & 56.36 & 56.59 & 64.71 & 78.79 & 16.13 & 69.06 & 58.91 \\
        SEALONG-14B              & 64.91 & 56.36 & 55.04 & 70.59 & 77.78 & 13.98 & 69.78 & 57.39 \\
        LoongRL-14B  & 64.91 & \bf 65.45 & \bf 65.89 & 70.59 & 80.81 & 15.05 & 70.50 & 61.46 \\
        RL-MemAgent-14B       & 68.42 & \bf 65.45 & 51.16 & 70.59 & 83.84 & 18.28 & 68.35 & 59.08 \\
        \textbf{\lib-14B} & \bf 75.44 & 61.82 & \bf 65.89 & \bf 76.47 & \bf 86.87 & \bf 20.43 & \bf 71.94 & \bf 64.52  \\
        \midrule

        Qwen3-8B & 66.67 & 54.55 & 55.81 & 52.94 & 85.86 & 18.28 & 66.19 & 58.23\\
        \bf \lib-8B & \bf 71.93 & \bf 63.64 & \bf 60.47 & \bf 58.82 & \bf 88.89 & \bf 20.43 & \bf 71.22 & \bf 62.82 \\

        \bottomrule
    \end{tabular}%
    }
    \vspace{-12px}
\end{table}